\documentclass[12pt,letterpaper]{report}
\usepackage{natbib}
\usepackage{geometry}
\usepackage{multirow}
\usepackage{fancyhdr}
\usepackage{afterpage}
\usepackage{graphicx}
\usepackage{amsmath,amssymb,amsbsy}
\usepackage{dcolumn,array}
\usepackage{tocloft}
\usepackage{asudis}
\usepackage{caption}
\usepackage{tabularx}
\usepackage{subcaption}
\usepackage{titlesec}
\usepackage{fancyhdr}
\usepackage{graphicx}
\usepackage{multirow}
\usepackage[table]{xcolor}
\usepackage[caption=false]{subfig}
\usepackage{amsmath}
\usepackage{tocloft}
\usepackage{booktabs}
\begin{document}
%-----------------------front matter
\fancypagestyle{contents}{
  \fancyhf{} % Clear header and footer
  \renewcommand{\headrulewidth}{0pt} % Remove header rule
  \fancyhead[L]{\textbf{TABLE OF CONTENTS}} % Header centered with TABLE OF CONTENTS
  \fancyfoot[C]{\thepage} % Footer centered with page number
}
\pagenumbering{roman}
\title{Multi-Modal Tumor Survival Prediction via Graph-Guided Mixture of Experts}
\author{Hirthik Mathavan}
\degreeName{Master of Science}
\paperType{Thesis}
\defensemonth{April}
\gradmonth{May}
\gradyear{2024}
\chair{Huan Liu}
\memberOne{Hasan Davulcu}
\memberTwo{YooJung Choi}
% \memberThree{}
% \memberFour{}

\maketitle
\doublespace
\begin{abstract}
Large Language Models (LLMs) have displayed impressive capabilities in handling tasks that require few demonstration examples, making them effective few-shot learners. Despite their potential, LLMs face challenges when it comes to addressing complex real-world tasks that involve multiple modalities or reasoning steps. For example, predicting cancer patients' survival period based on clinical data, cell slides, and genomics poses significant logistical complexities. Although several approaches have been proposed to tackle these challenges, they often fall short in achieving promising performance due to their inability to consider all modalities simultaneously or account for missing modalities, variations in modalities, and the integration of multi-modal data, ultimately compromising their effectiveness.

This thesis proposes a novel approach for multi-modal tumor survival prediction to address these limitations. Taking inspiration from recent advancements in LLMs, particularly Mixture of Experts (MoE)-based models, a graph-guided MoE framework is introduced. This framework utilizes a graph structure to manage the predictions effectively and combines multiple models to enhance predictive power. Rather than training a single foundation model for end-to-end survival prediction, the approach leverages a MOE-guided ensemble to manage model callings as tools automatically. By leveraging the strengths of existing models and guiding them through a MOE framework, the aim is to achieve better performance and more accurate predictions in complex real-world tasks. Experiments and analysis on the TCGA-LUAD dataset show improved performance over the individual modal and vanilla ensemble models.

\end{abstract}

\dedicationpage{This work is dedicated with love and gratitude to my loving parents, family, friends, and mentors who have always believed in me and supported me in all my endeavors.}
\begin{acknowledgements}
I want to express my deepest gratitude to my advisor, Dr. Huan Liu, for his unwavering support and guidance throughout my master's journey. His assistance allowed me to delve into research in this field. I also owe a debt of gratitude to Zhen Tan, whose mentorship has been invaluable every step of the way. This journey would not have been possible without his dedicated help.

Heartfelt thanks are due to all the members of the DMML lab group for their willingness to listen and share their insights. To my friends at ASU, your companionship has made this journey truly unforgettable. I would like to mention my brother, Jaanav Mathavan, and my parents, whose constant encouragement has been my driving force.

I am sincerely grateful to my thesis committee members, Dr. Hasan Davulcu and Dr. YooJung Choi, for their guidance and support. This work would not have been possible without the collective encouragement and assistance of everyone mentioned here. From the depths of my heart, thank you all.
\end{acknowledgements}
\tableofcontents
% This puts the word "Page" right justified above everything else.
\addtocontents{toc}{~\hfill Page\par}
% Asking LaTeX for a new page here guarantees that the LOF is on a separate page
% after the TOC ends.
\newpage
% Making the LOT and LOF "parts" rather than chapters gets them indented at
% level -1 according to the chart: top of page 4 of the document at
% ftp://tug.ctan.org/pub/tex-archive/macros/latex/contrib/tocloft/tocloft.pdf
\addcontentsline{toc}{part}{LIST OF TABLES}
\renewcommand{\cftlabel}{Table}
\listoftables
% This gets the headers for the LOT right on the first page.  Subsequent pages
% are handled by the fancyhdr code in the asudis.sty file.
\addtocontents{lot}{Table~\hfill Page \par}
\newpage
\addcontentsline{toc}{part}{LIST OF FIGURES}
\addtocontents{toc}{CHAPTER \par}
\renewcommand{\cftlabel}{Figure}
\listoffigures
% This gets the headers for the LOF right on the first page.  Subsequent pages
% are handled by the fancyhdr code in the asudis.sty file.
\addtocontents{lof}{Figure~\hfill Page \par}
%-----------------------body
\doublespace
\pagenumbering{arabic}
\chapter{INTRODUCTION}
\pagestyle{plain}
% 1. Graph is important (examples)
% 2. Reason: Relation (GNN)
% 3. The relation modeling can help with data-scarce issues [domain-specifc task].
% 4. SBP
% 5. LLM emerge as a new few-shot learner, but it fails in specific domains (AVIS)
% 6. Use graph to guide the inference of LLMs
% 7. Inspired by those recent LLMs, which are usually MoE-based model
% 8. graph-guided MoE (dense) - gating is like a fully-connected GAT model
% 9. MoE (Sparse) - GAT with graph condensation
% 10. Our domain - multi-modal tumor survival prediction [data-scarce]

Graphs play a crucial role in various research fields, including social network analysis \citep{SNA}, bioinformatics \citep{BINF}, recommendation systems \citep{ReSs}, and foundation models \citep{AVIS}. In the realm of sentiment analysis on social media, graphs model user interactions, allowing algorithms to classify sentiments based on structural and attribute features. Graphs are also used as traversals in Visual Information Seeking with Large Language Model Agents as transitions and in Foundation Models to tackle complex models and real-world problems \citep{AVIS}.

Graphs have found many applications due to the relation modeling within the graph-structured data. Graph Neural Networks (GNNs) have emerged as powerful tools for modeling relationships and dependencies within graph-structured data. Unlike traditional neural networks operating on grid-like data (e.g., images, sequences), GNNs exploit the inherent graph structure to capture local and global interactions among nodes. GNNs can effectively encode complex relational information into node embeddings, enabling tasks such as node classification, link prediction, and graph generation by aggregating information from neighboring nodes through message-passing mechanisms. The relational modeling of graphs holds significant promise in mitigating data scarcity issues, particularly in domain-specific tasks such as in Healthcare \citep{GE2023104458}, Social Sciences \citep{SociS}, Cyber Security \citep{CyberS}, etc, where labeled data may be limited or expensive. To understand K-shot learning and Graphs, we explored Few-shot learning, a subset of machine learning that has gained popularity for constructing models with minimal labeled data. We focused on meta-learning, a powerful technique enabling rapid adaptation to new tasks with minimal labeled data. Addressing the underexplored realm of inductive few-shot learning in graph node classification, our research underscored the necessity for models to adapt to novel contexts and explored practical baseline approaches to enhance inductive few-shot node classification tasks \citep{[ILP]}.
% Meta-learning, or learning to learn, is a powerful technique for few-shot learning, where the model is trained on various tasks for rapid adaptation to new tasks with limited labeled data. In graph node classification, meta-learning mainly explores the transductive setting, where the model is trained and evaluated on the same graph. However, the inductive setting, where the model is tested on new, unseen graphs, poses additional challenges, particularly with the message-passing mechanism in graph neural networks (GNNs). Our research addressed the underexplored realm of inductive, few-shot learning in graph node classification, acknowledging the evolving nature of social media platforms and user dynamics. We emphasized the need for models to adapt to novel contexts and empirically showed the limitations of existing meta-learning frameworks in the inductive setting. We proposed a straightforward yet practical baseline approach to bridge the gap and pave the way for enhanced inductive few-shot node classification tasks . 

Now, LLMs have impressive capabilities in solving tasks with few demonstration examples emerging as a new few-shot learner. For example, recent work with Visual Information tasks has been tackled with LLM Agents using graphs to guide inferences \citep{AVIS}. However, they still face limitations in handling complex real-world tasks involving multiple modalities, which are often logistically complicated and require dealing with multi-modal data. For instance, predicting the survival period of cancer patients based on clinical data, cell slides, and symptom descriptions is a challenging task, even for experienced doctors. Several existing works have attempted to solve these challenges, but they either fail to consider all modalities simultaneously or cannot achieve promising performance.

This thesis proposes a novel approach to address the issue of multi-modal tumor survival prediction. We are inspired by recent LLMs (MoE-based models) and have developed a graph-guided MoE that uses a graph structure to manage the callings of existing predictions. Rather than directly training a single foundation model to perform end-to-end survival prediction, our method can be viewed as a MOE-guided ensemble that automatically manages the callings for models as tools. We leverage the strengths of existing models and combine them to enhance their predictive power. By using a MOE-guided ensemble, we manage the callings of existing models more effectively and efficiently, leading to better performance and more accurate predictions. Our goal is to leverage the strengths of multiple models and guide them to achieve better performance and more accurate predictions in complex real-world tasks.

The summary of our contribution is as follows,
\begin{itemize}
    \item \textbf{New Framework:} The thesis presents a novel approach for multi-modal tumor survival prediction, addressing the limitations of existing models by introducing a graph-guided Mixture of Experts framework. 

\item \textbf{Improved Performance:} Experiments and analysis on the TCGA LUAD dataset demonstrate that the proposed approach performs better than other survival prediction models. We evaluated six models, including the proposed framework. 

\item \textbf{Practical Relevance:} The proposed approach has the potential to translate into practical clinical applications, with implications for improved patient care and personalized treatment strategies in oncology. This contribution aligns with the broader goal of advancing predictive modeling techniques to benefit healthcare and medical practitioners.
\end{itemize}

\chapter{RELATED WORK \& BACKGROUND}

% Background: 
% Few-shot/Zero Shot (Meta-learning, Contrastive learning)
% Inductive / transductive

\noindent
\section{Graphs}
\subsection{Overview, Applications, and Representation}
% \textbf{What is a graph?}

A graph is a versatile data structure that encapsulates relationships between objects, where edges connect pairs of objects, and the connected objects are referred to as vertices or nodes. With applications spanning social networks, molecular structures, citation networks, and knowledge graphs, graphs serve as a fundamental model for representing diverse relationships. Graphs are utilized for many tasks, ranging from graph-level predictions to node and edge-level property predictions and even sub-graph tasks like community detection. Representations of graphs commonly include sets of edges and nodes or adjacency matrices, reflecting the intricate connectivity among entities. Despite their complexity, graphs offer powerful insights into relational data, catering to various analytical and predictive tasks in various domains.

\subsection{Graph Neural Networks}

Graph Neural Networks (GNNs) are a class of neural networks designed to operate on graph-structured data. They have become increasingly popular due to their ability to model and learn from graph-structured data effectively. GNNs typically consist of successive layers, and each layer represents a node as the combination (aggregation) of the representations of its neighbors and itself from the previous layer (message passing). GNNs adhere to certain requirements, such as permutation invariance and equivariance, which means that the network's representation of a graph and its permutations should be the same after going through the network. There are various ways to aggregate messages from neighboring nodes in GNNs, such as summing, averaging, or using attention mechanisms to weigh the importance of neighboring nodes. Aggregation techniques play a crucial role in capturing the graph's structural information and generating meaningful representations for nodes within the graph.
\noindent
% \textbf{Few-shot Learning:}
\section{Few-shot Learning}

Few-Shot Learning (FSL) is a machine learning framework that addresses the challenge of training models with only a few labeled samples per class, enabling them to generalize over new categories of data not seen during training. This framework falls under the larger paradigm of meta-learning, which involves learning to learn. FSL aims to mimic human ability, as humans can identify new classes of data using only a few examples and our previously learned knowledge.

% \subsection{Key terms related to Few-Shot Learning}

% \textbf{Support Set:} This consists of the few labeled samples per novel category of data that a pre-trained model will use to generalize on these new classes.

% \textbf{Query Set:} This consists of the samples from the new and old categories of data on which the model needs to generalize using previous knowledge and information gained from the support set.

% \textbf{N-way K-shot learning scheme:} This term is common in the FSL literature and describes the few-shot problem statement that a model will be dealing with. "N-way" indicates the number of novel categories on which a pre-trained model needs to generalize, and "K-shot" defines the number of labeled samples available in the support set for each of the "N" novel classes.

% \subsection{Why Few-Shot Learning?}
\begin{figure}[htp]
    \centering
    \includegraphics[width=14cm]{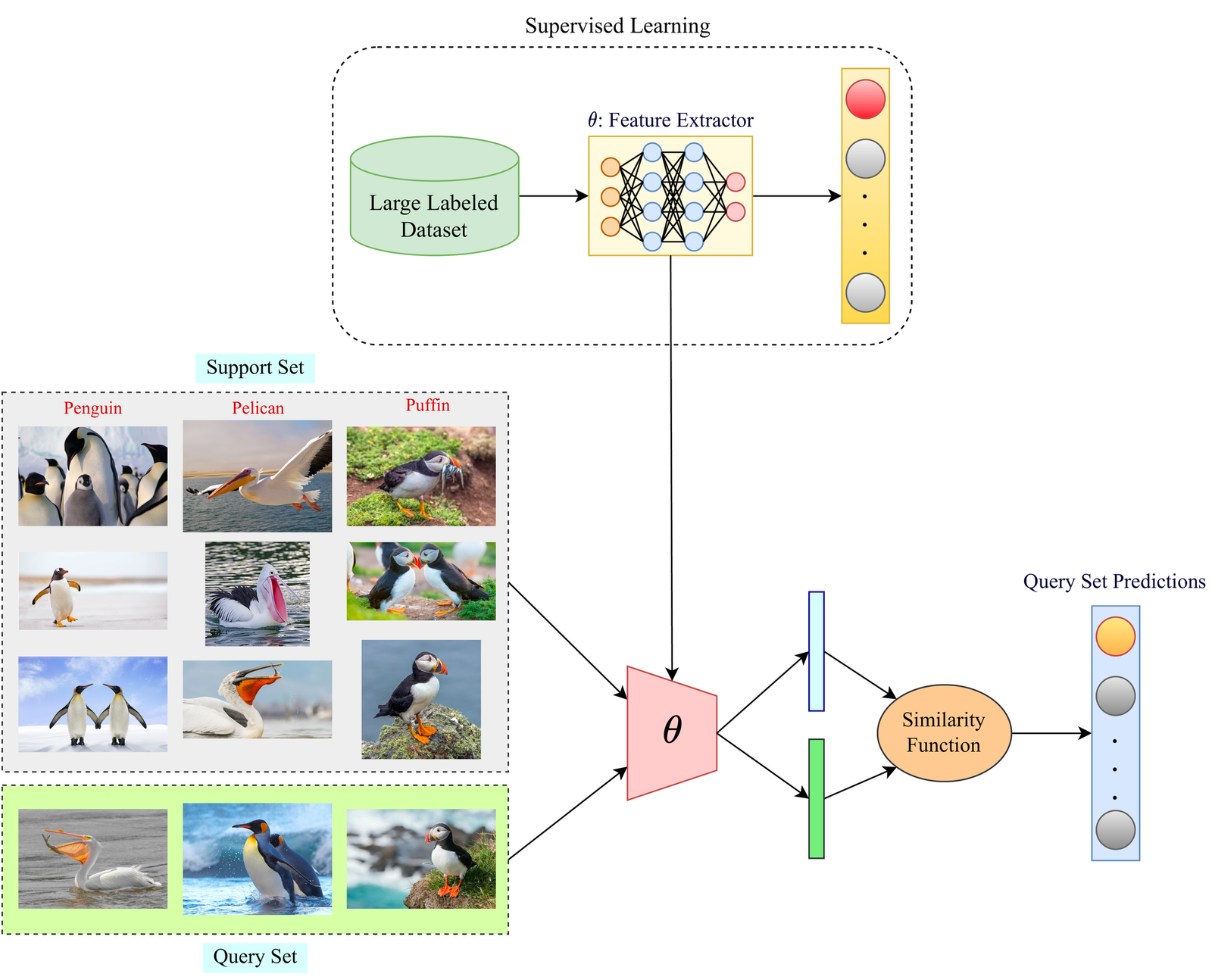}
    \captionsetup{justification=centering}
    \caption{General Few Shot Learning Approach
 }
    \label{fig:galaxy}
\end{figure}
Traditional supervised learning methods use large quantities of labeled data for training, and the test set comprises data samples that must not only belong to the same categories as the training set but also come from a similar statistical distribution. Few-shot learning alleviates the problems associated with traditional supervised learning in the following ways:
\begin{itemize}
  \item Eliminates the requirement for large volumes of costly labeled data.
  \item Enables the extension of pre-trained models to new data categories without re-training from scratch.
  \item Enables learning about rare categories of data with exposure to only limited prior information.
  \item Allows models to be extended to different data domains as long as the data in the support and query sets are coherent.
\end{itemize}

Traditional approaches in few-shot learning include meta-learning, metric learning, and data augmentation. Meta-learning, also known as learning to learn, involves training models on various tasks to rapidly adapt to new tasks with limited labeled data. Popular meta-learning algorithms include MAML (Model-Agnostic Meta-Learning) \citep{[1]} and Prototypical Networks \citep{[Proto]}, which have demonstrated success in few-shot classification tasks. Our paper \citep{[ILP]} introduced a competitive baseline approach for few-shot node classification under the inductive settings. 

Metric learning approaches focus on learning a distance metric that measures similarity between samples. Siamese networks \citep{Koch2015SiameseNN} and triplet loss \citep{Schroff2015FaceNetAU} are common techniques used in metric learning for few-shot learning tasks. Data augmentation techniques such as generative models (e.g., GANs) and synthetic data generation have also been explored to augment the training data and improve model generalization in few-shot scenarios, \citep{FSN}. 
% \noindent

\textbf{Zero-shot Learning:}
% \section{Zero-shot Learning}

On the other hand, zero-shot learning (ZSL) is a machine learning scenario where an AI model is trained to recognize and categorize objects or concepts without seeing any examples of those categories or concepts beforehand. In ZSL, the model is not trained on any labeled examples of the unseen classes, and it is asked to make predictions post-training. Instead, ZSL uses auxiliary information such as textual descriptions, attributes, embedded representations, or other semantic information relevant to the task. It typically outputs a probability vector representing the likelihood that a given input belongs to certain classes. ZSL has become notable in computer vision and natural language processing, as it enables AI models to generalize quickly to many semantic categories with minimal training overhead, particularly in scenarios where labeled examples are scarce or non-existent. Attribute-based methods represent classes using attribute vectors describing various semantic attributes (e.g., color, shape) and learn a compatibility function to predict class labels based on attribute similarity \citep{inproceedings}.

Semantic embedding approaches map classes and features into a common semantic space, enabling inference for unseen classes based on their similarity to seen classes \citep{NIPS2013_7cce53cf}. Text-based zero-shot learning methods leverage textual descriptions or natural language queries to infer class labels for unseen examples. Techniques such as natural language embeddings and text-image matching have been employed for zero-shot recognition, \citep{NIPS2013_2d6cc4b2}.

% \noindent
% \textbf{Foundation Models:}

\section{Large Language Models}

Foundation models are large deep-learning neural networks that have revolutionized the field of machine learning. These models are trained on extensive datasets and are designed to serve as a starting point for developing machine learning (ML) models that power new applications more efficiently and cost-effectively. Essentially, the "foundation model" refers to ML models trained on a broad spectrum of generalized and unlabeled data. They can perform various general tasks, such as understanding language, generating text and images, and conversing in natural language.

% % \textbf{Unique Aspects of Foundation Models:}
% \subsection{Unique Aspects of Foundation Models}
% \begin{itemize}
% \item \textbf{Adaptability: }Foundation models can perform a wide range of disparate tasks with a high degree of accuracy based on input prompts. Tasks include natural language processing (NLP), question answering, and image classification. They are different from traditional ML models as they are general-purpose and can be used as base models for developing more specialized downstream applications.

% \item \textbf{Evolution in Size and Complexity: }The size and general-purpose nature of foundation models distinguish them from traditional models. They have increased in size and complexity over the years. For example, GPT-4, released in 2023, was trained using 170 trillion parameters and a 45 GB training dataset, representing a significant leap from previous models.
% \end{itemize}

% % \textbf{Functionality of Foundation Models:}
% \subsection{Functionality of Foundation Models}
% \begin{itemize}
% \item \textbf{Generative Artificial Intelligence (AI):} Foundation models are a form of generative AI that uses learned patterns and relationships to predict the next item in a sequence, creating outputs from one or more input prompts in human language instructions.

% \item \textbf{Continuous Learning:} Even though pre-trained, foundation models can continue to learn from data inputs or prompts during inference, enabling the development of comprehensive outputs through carefully curated prompts.
% \end{itemize}

Foundation models, particularly Large Language Models (LLMs), have revolutionized various fields by exhibiting remarkable abilities in solving complex tasks with minimal demonstration examples. These models are the cornerstone for numerous downstream applications and have spurred research in diverse areas.

\textbf{Large Language Models (LLMs):}
LLMs, such as OpenAI's GPT series and Google's BERT, have demonstrated exceptional capabilities in understanding and generating human-like text. These models are trained on vast amounts of text data and can be fine-tuned for specific tasks, ranging from language understanding and generation to translation and summarization. Recent advancements in LLMs have also enabled them to tackle multimodal tasks by integrating textual and non-textual modalities, expanding their applicability to a broader range of tasks and domains.

\textbf{Mixture of Experts (MoE):}
Mixture of Experts is a powerful neural network architecture that combines multiple expert models within a gating network to handle diverse tasks or modalities. Each expert specializes in a specific sub-task or domain, while the gating network dynamically selects the most relevant expert for a given input. MoE has been applied successfully in various domains, including natural language processing (GPT-4), computer vision, and recommendation systems to improve model performance and robustness \citep{shazeer2017}.
\begin{figure}[htp]
    \centering
    \includegraphics[width=14cm]{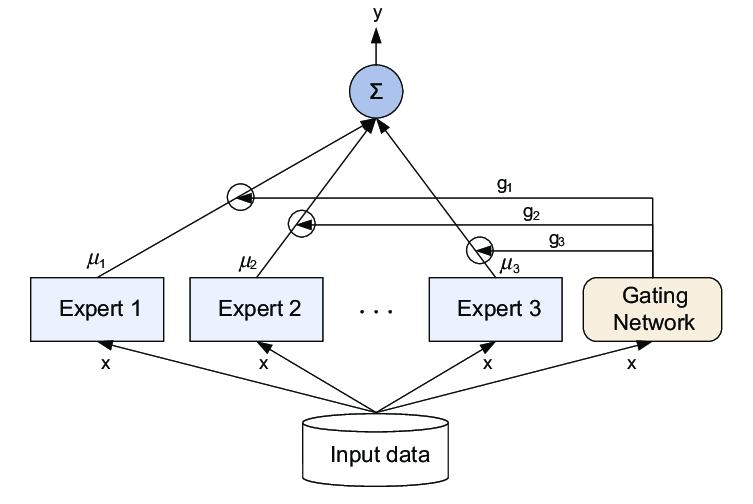}
    \captionsetup{justification=centering}
    \caption{General Mixture of Experts Approach
 }
    \label{fig:galaxy}
\end{figure}

\textbf{LLM for Multi-modal Tasks:}

LLMs have been extended to handle multi-modal tasks by incorporating both textual and non-textual inputs, such as images, audio, or structured data. This extension enables LLMs to understand and generate content across multiple modalities, opening up new opportunities for applications in areas like image captioning, video understanding, and multimodal translation.

\textbf{a. ChatGPT (GPT-3.5):}
GPT-3.5 is a large language model developed by OpenAI. GPT-3.5 is renowned for its state-of-the-art performance in various language understanding and generation tasks. Built upon a transformer architecture, GPT-3.5 comprises an expansive neural network with 175 billion parameters, enabling it to capture intricate linguistic patterns and nuances across diverse contexts.

The methodology underlying GPT-3.5 involves pre-training on vast amounts of text data from the internet, encompassing many languages and domains. The model learns to predict the next token in a sequence given the preceding context through unsupervised learning, thereby acquiring a rich understanding of language structure and semantics. Moreover, GPT-3.5 employs a multi-layer architecture with attention mechanisms, facilitating the capture of long-range dependencies and contextual information within text sequences.

\textbf{b. Multimodal LLM:}
Multimodal Large Language Models (LLMs) represent a sophisticated approach to processing and understanding diverse forms of information, seamlessly integrating textual and visual data. They are adept at generating captions for images, answering questions about visual content, and analyzing sentiment across multiple modalities. Leveraging pre-trained LLMs as their foundation, these models incorporate specialized layers or modules designed to handle and fuse the complexities of multimodal inputs, enabling them to extract richer insights and produce more nuanced outputs across various tasks and domains.
% Multimodal LLMs integrate textual and visual information to perform tasks such as image captioning, visual question answering, and multimodal sentiment analysis. These models leverage pre-trained LLMs as the backbone and incorporate additional layers or modules to effectively process and fuse multimodal inputs.

\textbf{c. Decision Making with LLM as an Agent:}
There has also been a surge of interest in applying
Large Language Models (LLMs) as autonomous agents. These agents are capable of interacting with external environments, making dynamic decisions based on real-time feedback, and consequently achieving specific goals. For example, Autonomous Visual Information Seeking with Large Language Model Agent (AVIS) \citep{AVIS}, method leverages a Large Language Model (LLM) to strategize the utilization of external tools dynamically and to investigate their outputs, thereby acquiring the indispensable knowledge needed to provide answers to the posed questions. We concentrate on this approach of decision-making in selecting appropriate tools/models for our problem (Cancer Survival Prediction). 

% \noindent
% \textbf{Cancer Survival Prediction:}

\section{Cancer Survival Prediction}

\subsection{Cancer Prognosis}
Globally, cancer-related deaths currently stand at approximately 10 million annually, with projections indicating that cancer will emerge as the primary cause of death in all countries during the 21st century \citep{1Bray2018}. In oncology, the prediction of time-to-event outcomes, such as cancer recurrence or mortality, plays a pivotal role in clinical decision-making. Survival analysis, specifically, is highly valued by patients, clinicians, researchers, and policymakers alike \citep{2Mariotto2014}.

Traditional cancer prognosis methods typically rely on population-level estimates tailored to specific cancer types and stages. However, these approaches overlook individual patient variations, including critical factors such as age at diagnosis. To address this limitation, several patient-specific prediction techniques have been introduced in clinical practice, integrating clinical data with laboratory measurements of validated biomarkers \citep{3Simmons2017}.

Despite these advancements, survival prediction often remains reliant on subjective interpretation and intuition \citep{4Hui2019} by clinicians, which can compromise accuracy and reproducibility \citep{5Cheon2016}. Thus, there is a pressing need for more objective and precise methodologies in cancer prognosis.
\subsection{Survival analysis}
Survival analysis is a field focused on understanding the time it takes for a particular event to occur. In the context of cancer, this typically refers to the duration between diagnosis and death caused by the disease. While the ideal scenario involves observing every patient until the event of interest, in reality, some patients may drop out of follow-up early, and in some cases, the event may not occur at all if the patient succumbs to another cause. When the event is not observed, the last known contact time is termed as censoring time. Despite not directly observing the event, censored observations still provide valuable information for modeling, as they offer a lower bound on the patient's survival time.

The semi-parametric Cox proportional hazards (CPH) model \citep{6Cox1972} is a commonly used statistical approach for analyzing survival data with censored observations \citep{7LeCun2015}. However, it has notable limitations. Firstly, it relies on a linear model, which limits its ability to capture non-linear relationships between input data and the risk of death. Additionally, the CPH model assumes that the effects of patient features remain constant over time, resulting in proportional predictions for patients at all follow-up time points. 

Initially, the integration of DL models within the CPH framework began with the utilization of a basic feed-forward neural network with a unimodal data input \citep{10Norgeot2019}. However, as the era of big data in precision medicine unfolds, there's a surge in the availability of diverse data modalities in routine clinical practice. This proliferation of big data necessitates robust modeling approaches, thereby underscoring the importance of DL-based methods. Consequently, recent studies have capitalized on modern DL techniques to enhance the capabilities of DL-based CPH models \citep{12Katzman2018}.

Prominent examples include methodologies tailored to leverage clinical and gene expression data, such as DeepSurv \citep{13Ching2018} and Cox-nnet \citep{14Lu2019}. Additionally, there are approaches targeting imaging data, such as CXR-risk for chest radiographs\citep{15Mukherjee2020}, LungNet\citep{16Zhang2020} for computed tomography (CT) images in lung cancer prognosis, and a model\citep{17Zhong2020} for gastric cancer survival prediction based on CT images. 

\subsection{Multimodal Cancer Data}
The availability of vast and varied datasets, encompassing clinical records, imaging scans, and molecular profiles, underscores the importance of integrating these data types using deep multimodal representation learning techniques \citep{20Baltrusaitis2019,21Yousefi2017}. Recent advancements have extended deep learning-based Cox survival models to incorporate diverse data modalities, enhancing predictive accuracy. For instance, SurvivalNet integrates various high-throughput molecular data modalities across different cancer types \citep{22Mobadersany2018}. The GSCNN system merges digital pathology images with validated genomic biomarkers in glioma patients \citep{23Huang2019}. SALMON tackles breast cancer by combining gene and microRNA expression data with clinical parameters and validated biomarkers \citep{24Cheerla2019}. 
% \rule{\textwidth}{2mm}
% \begin{table}[htbp]
% \centering
% \caption{Summary of Covariates for Patients (LUAD) }
% \scalebox{0.7}{
% \begin{tabular}{lcc}
% \toprule
% \textbf{Covariates} & \textbf{Group} & \textbf{Patients (N = 522)} \\
% \midrule
% Survival time & & $902.51 \pm 892.15$ \\
% \midrule
% \multirow{2}{*}{Vital status} & Alive & 334 (63.98\%) \\
% & Dead & 188 (36.02\%) \\
% \midrule
% \multirow{4}{*}{Stage} & I & 280 (53.64\%) \\
% & II & 130 (24.90\%) \\
% & III & 86 (16.48\%) \\
% & IV & 26 (4.98\%) \\
% \midrule
% \multirow{4}{*}{T stage} & T1 & 172 (32.95\%) \\
% & T2 & 281 (53.83\%) \\
% & T3 & 47 (9.00\%) \\
% & T4 & 22 (4.21\%) \\
% \midrule
% \multirow{4}{*}{N stage} & N0 & 342 (65.52\%) \\
% & N1 & 99 (18.97\%) \\
% & N2 & 75 (14.37\%) \\
% & N3 & 6 (1.15\%) \\
% \midrule
% \multirow{2}{*}{M stage} & M0 & 496 (95.02\%) \\
% & M1 & 26 (4.98\%) \\
% \midrule
% \multirow{2}{*}{Age} & $\leq65$ & 250 (47.89\%) \\
% & $>65$ & 272 (52.11\%) \\
% \midrule
% \multirow{2}{*}{Gender} & Female & 280 (53.64\%) \\
% & Male & 242 (46.36\%) \\
% \bottomrule
% \end{tabular}}
% \end{table}
% \usepackage{multirow}
\begin{table}[]
\scriptsize
\centering
\caption{Summary of Covariates for Patients (LUAD)}
\begin{tabular}{ccc}
\hline
\rowcolor{gray!30}
\textbf{Covariates}           & \textbf{Group}  & \textbf{Patients (N = 522)} \\ \hline
Survival time                 &                 & 902.51 ± 892.15             \\ \hline
\multirow{2}{*}{Vital status} & Alive           & 334 (63.98\%)               \\ \cline{2-3} 
                              & Dead            & 188  (36.02\%)              \\ \hline
\multirow{4}{*}{Stage}        & I               & 280 (53.64\%)               \\ \cline{2-3} 
                              & II              & 130 (24.90\%)               \\ \cline{2-3} 
                              & III             & 86 (16.48\%)                \\ \cline{2-3} 
                              & IV              & 26 (4.98\%)                 \\ \hline
\multirow{4}{*}{T stage}      & T1              & 172 (32.95\%)               \\ \cline{2-3} 
                              & T2              & 281 (53.83\%)               \\ \cline{2-3} 
                              & T3              & 47 (9.00\%)                 \\ \cline{2-3} 
                              & T4              & 22 (4.21\%)                 \\ \hline
\multirow{4}{*}{N stage}      & N0              & 342 (65.52\%)               \\ \cline{2-3} 
                              & N1              & 99 (18.97\%)                \\ \cline{2-3} 
                              & N2              & 75 (14.37\%)                \\ \cline{2-3} 
                              & N3              & 6 (1.15\%)                  \\ \hline
\multirow{2}{*}{M stage}      & M0              & 496 (95.02\%)               \\ \cline{2-3} 
                              & M1              & 26 (4.98\%)                 \\ \hline
\multirow{2}{*}{Age}          & $\le$ 65            & 250 (47.89\%)               \\ \cline{2-3} 
                              & \textgreater 65 & 272 (52.11\%)               \\ \hline
\multirow{2}{*}{Gender}       & Female          & 280 (53.64\%)               \\ \cline{2-3} 
                              & Male            & 242 (46.36\%)               \\ \hline
\end{tabular}
\end{table}

% \begin{table}[htp]
% \scriptsize
% \centering
% \caption{Summary of Covariates for Patients (LUAD)}
% \small
% % \setlength{\extrarowheight}{-6pt} % Adjust vertical spacing
% \begin{tabularx}{0.6\textwidth}{lcc}
% \toprule
% \textbf{Covariates} & \textbf{Group} & \textbf{Patients (N = 522)} \\
% \midrule
% Survival time & & $902.51 \pm 892.15$ \\
% \midrule
% \multirow{2}{*}{Vital status} & Alive & 334 (63.98\%) \\
% & Dead & 188 (36.02\%) \\
% \midrule
% \multirow{4}{*}{Stage} & I & 280 (53.64\%) \\
% & II & 130 (24.90\%) \\
% & III & 86 (16.48\%) \\
% & IV & 26 (4.98\%) \\
% \midrule
% \multirow{4}{*}{T stage} & T1 & 172 (32.95\%) \\
% & T2 & 281 (53.83\%) \\
% & T3 & 47 (9.00\%) \\
% & T4 & 22 (4.21\%) \\
% \midrule
% \multirow{4}{*}{N stage} & N0 & 342 (65.52\%) \\
% & N1 & 99 (18.97\%) \\
% & N2 & 75 (14.37\%) \\
% & N3 & 6 (1.15\%) \\
% \midrule
% \multirow{2}{*}{M stage} & M0 & 496 (95.02\%) \\
% & M1 & 26 (4.98\%) \\
% \midrule
% \multirow{2}{*}{Age} & $\leq65$ & 250 (47.89\%) \\
% & $>65$ & 272 (52.11\%) \\
% \midrule
% \multirow{2}{*}{Gender} & Female & 280 (53.64\%) \\
% & Male & 242 (46.36\%) \\
% \bottomrule
% \end{tabularx}
% \end{table}
% \begin{figure}[htp]
%     \centering
%     \includegraphics[width=14cm]{Clinical-pathological-data-of-lung-adenocarcinoma-LUAD-from-The-Cancer-Genome-Atlas.png}
%     % \includegraphics[width=8cm]{Lasso.pdf}
%     \captionsetup{justification=centering}
%     \caption{Clinical-pathological-data-of-lung-adenocarcinoma-LUAD-from-The-Cancer-Genome-Atlas
%  }
%     \label{fig:galaxy}
% \end{figure}

\begin{figure}[htp]
    \centering
    \includegraphics[width=14cm]{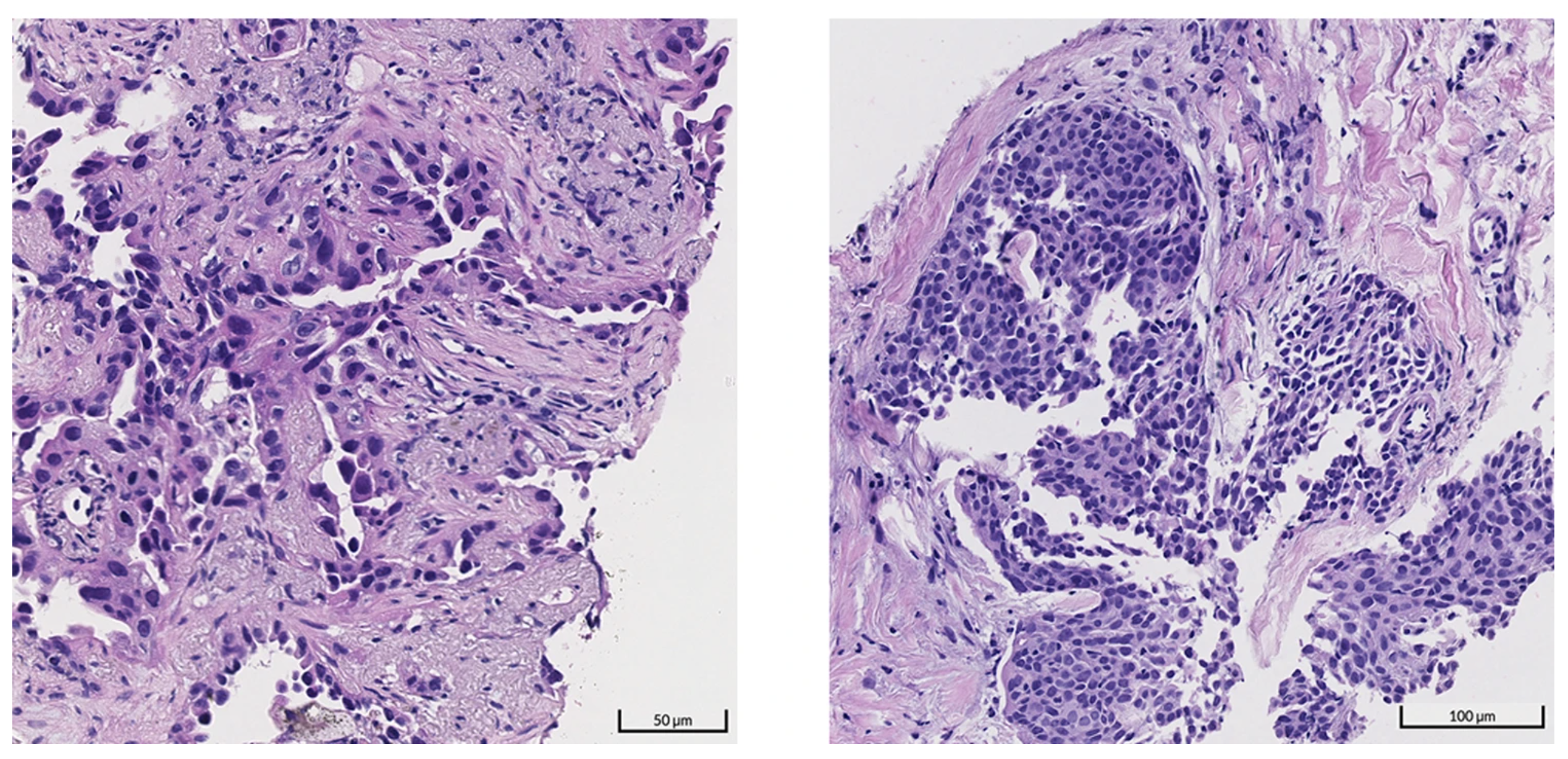}
    \captionsetup{justification=centering}
    \caption{Tissue Slides of LUAD Data
 }
    \label{fig:galaxy}
\end{figure}

\textbf{Cancer survival prediction} is a critical task in healthcare, aiming to estimate the likelihood of a patient's survival based on various clinical and demographic factors. Machine learning techniques, including deep learning and ensemble methods, have been employed to develop predictive models for cancer survival prognosis.

Other researches in cancer survival prediction has focused on leveraging diverse data sources, including clinical records, genomic data, imaging scans, and patient demographics, to build accurate and reliable prediction models. Feature selection, dimensionality reduction, and model interpretability techniques have been explored to enhance model performance and clinical interpretability. For e.g. \citep{10061470} addresses the challenge of multimodal cancer survival prediction by leveraging related pathological, clinical, and genomic features. It integrates graph convolutional networks (GCNs) and a hypergraph convolutional network (HCN) to capture intra- and inter-modal interactions within flexible and interpretable multimodal graphs. \citep{WANG2022102559} proposes a novel self-supervised learning strategy, called semantically-relevant contrastive learning (SRCL), tailored for histopathological image analysis, leveraging a hybrid model (CTransPath) pretrained on unlabeled data to generate informative representations, achieving state-of-the-art performance across various downstream tasks.

% Foundation Models: 
% LLMs
% Mixture of Experts
% LLM for multi-modal tasks: (a) MLLM (b) Tool planning

% cancer survival prediction

% For related work:
% This is where I include our SBP paper... what did we do there and its results,
% AVIS paper 
% and one-two more papers related to cancer survival prediction.

\chapter{METHODOLOGY}
% The proposed approach to addressing the issue of multimodal tumor survival prediction is a Mixture of expert-based model that utilize multiple existing survival prediction models that work with different modalities and act as experts within the MoE framework. 
% In a typical ensemble scenario, all models are trained on the same dataset, and their outputs are combined through simple averaging, weighted mean, or majority voting. However, in a mixture of experts, each “expert” model within the ensemble is only trained on a subset of data where it can achieve optimal performance, thus narrowing the model’s focus. The goal is to leverage the strengths of multiple models and guide them to achieve better performance and more accurate predictions in complex real-world tasks.
\section{Preliminary}
\subsection{Prompting ChatGPT-3.5}
We employ the GPT-3.5 language model, developed by OpenAI, as a baseline. In our research, we exploit GPT-3.5's capabilities through zero-shot learning to determine which expert model to use for a particular patient's tumor data and see if it can choose a model for us.

% \section{ChatGPT (3.5)}
% \begin{figure}[htp]
%     \centering
%     \includegraphics[width=8cm]{Ex1_a.png}
%     \includegraphics[width=7cm]{Ex1_b.png}
%     \captionsetup{justification=centering}
%     \caption{Example 1}
%     \label{fig:galaxy}
% \end{figure}
% \begin{figure}[htp]
%     \centering
%     \includegraphics[width=8cm]{Ex2_b.png}
%     \includegraphics[width=7cm]{Ex2_a.png}
%     \captionsetup{justification=centering}
%     \caption{Example 2}
%     \label{fig:galaxy}
% \end{figure}
% \begin{figure}[htp]
%     \centering
%     \includegraphics[width=8cm]{Ex3_b.png}
%     \includegraphics[width=7cm]{Ex3_a.png}
%     \captionsetup{justification=centering}
%     \caption{Example 3}
%     \label{fig:galaxy}
% \end{figure}
% \begin{figure}[htp]
%     \centering
%     \includegraphics[width=8cm]{Ex4_b.png}
%     \includegraphics[width=7cm]{Ex4_a.png}
%     \captionsetup{justification=centering}
%     \caption{Example 4}
%     \label{fig:galaxy}
% \end{figure}

\begin{figure}[htbp]
    \centering
    \begin{subfigure}{0.49\textwidth}
        \centering
        \includegraphics[width=\linewidth]{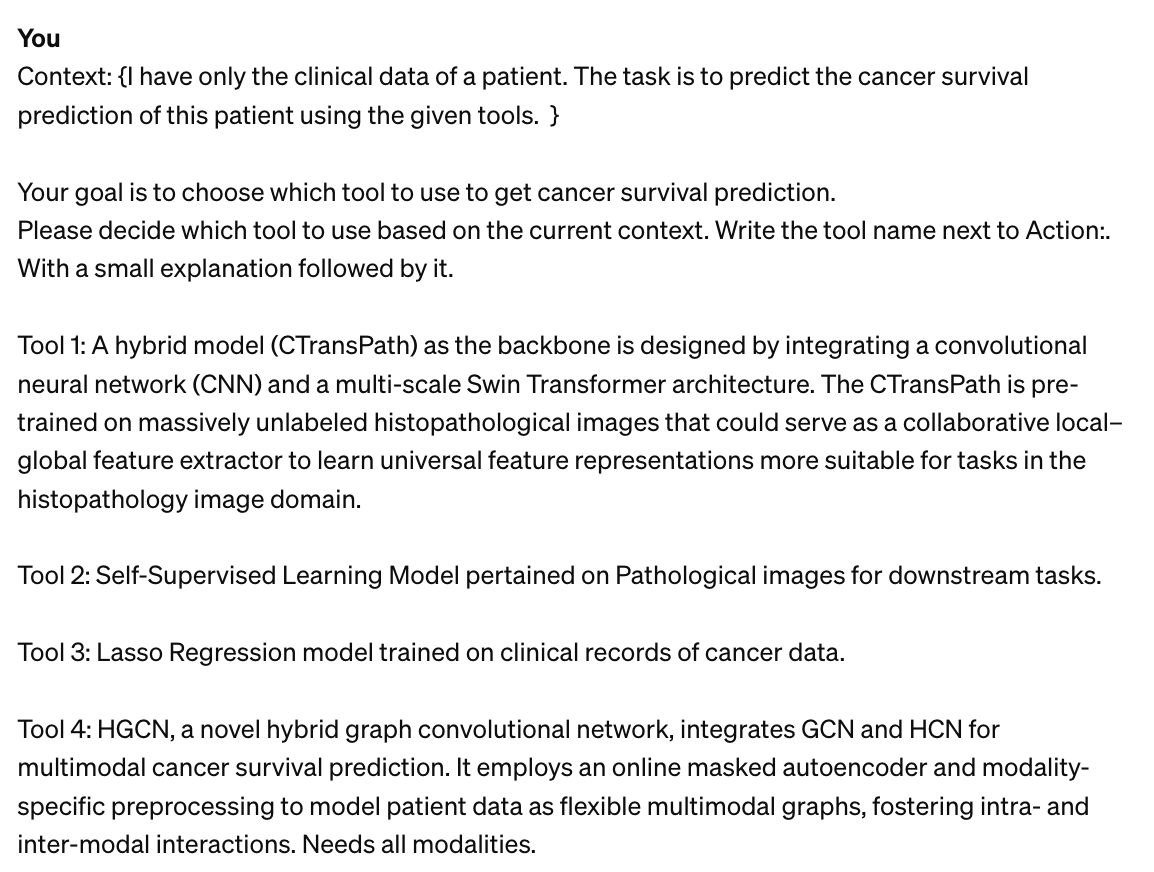}
        % \caption{Caption for image 1}
        \label{fig:sub1}
    \end{subfigure}
    \hfill
    \begin{subfigure}{0.49\textwidth}
        \centering
        \includegraphics[width=\linewidth]{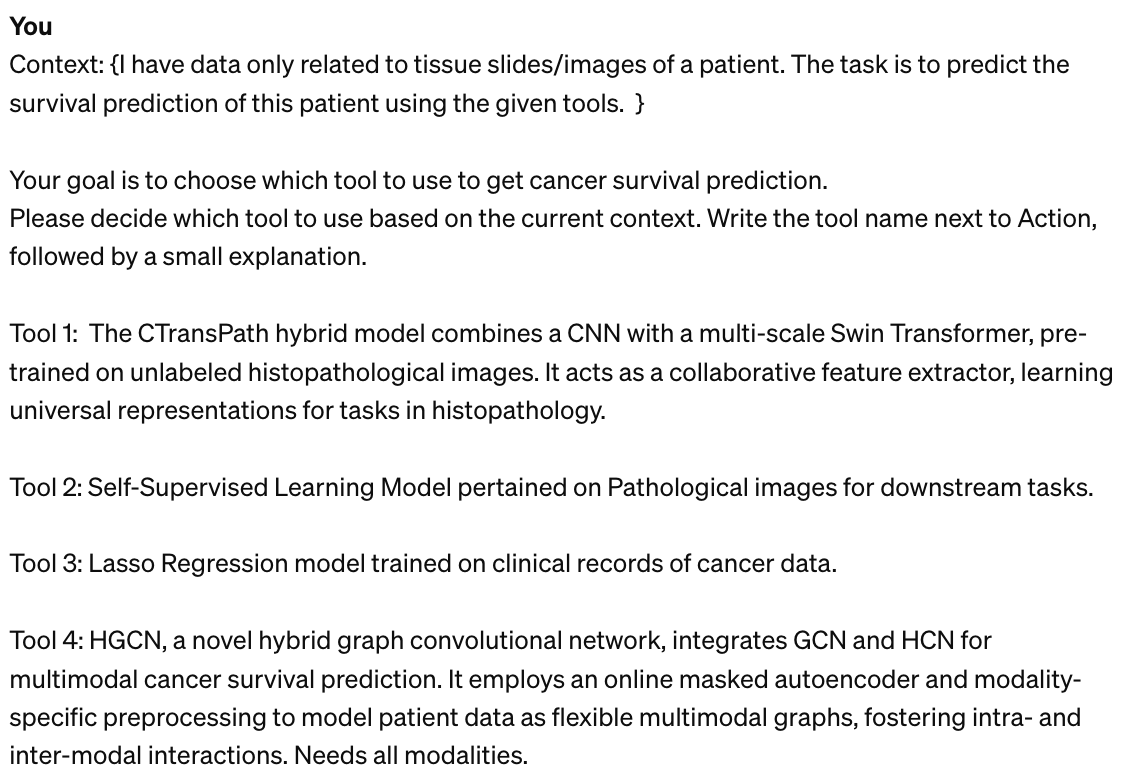}
        % \caption{Caption for image 2}
        \label{fig:sub2}
    \end{subfigure}
    \begin{subfigure}{0.49\textwidth}
        \centering
        \includegraphics[width=\linewidth]{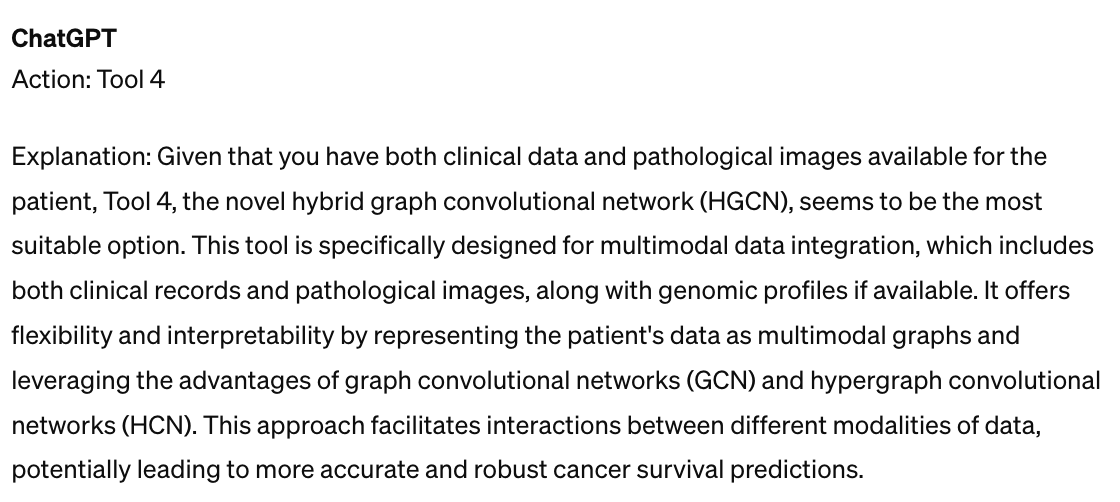}
        \caption{Example 1}
        \label{fig:sub3}
    \end{subfigure}
    \hfill
    \begin{subfigure}{0.49\textwidth}
        \centering
        \includegraphics[width=\linewidth]{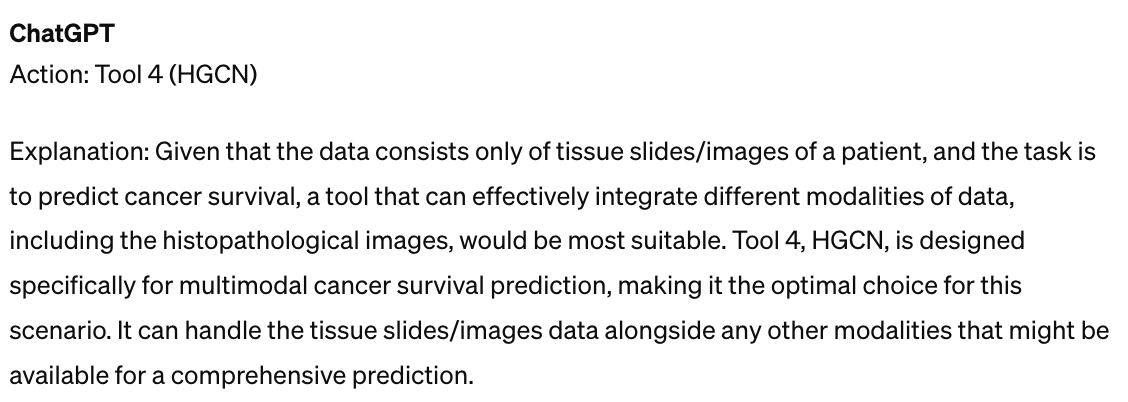}
        \caption{Example 2}
        \label{fig:sub4}
    \end{subfigure}
    % \caption{Main caption for the entire figure}
    \label{fig:main}
\end{figure}

\begin{figure}[htbp]
    \centering
    \begin{subfigure}{0.49\textwidth}
        \centering
        \includegraphics[width=\linewidth]{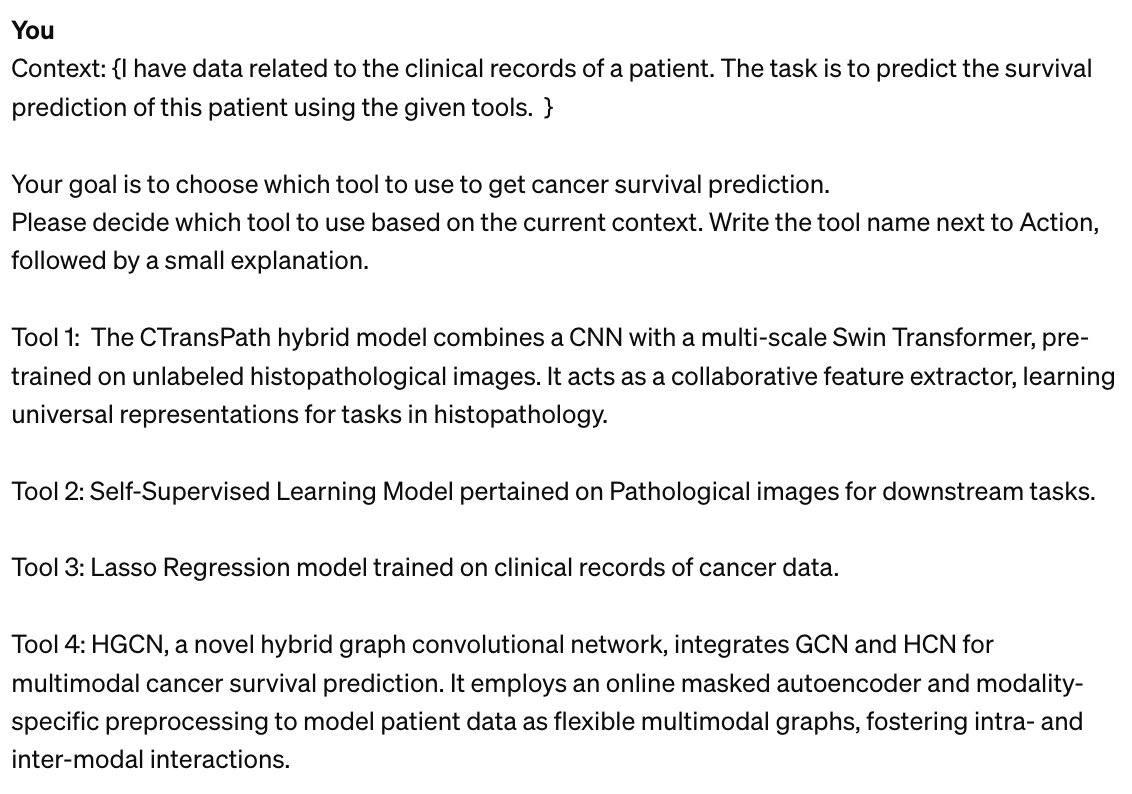}
        % \caption{Caption for image 1}
        \label{fig:sub1}
    \end{subfigure}
    \hfill
    \begin{subfigure}{0.49\textwidth}
        \centering
        \includegraphics[width=\linewidth]{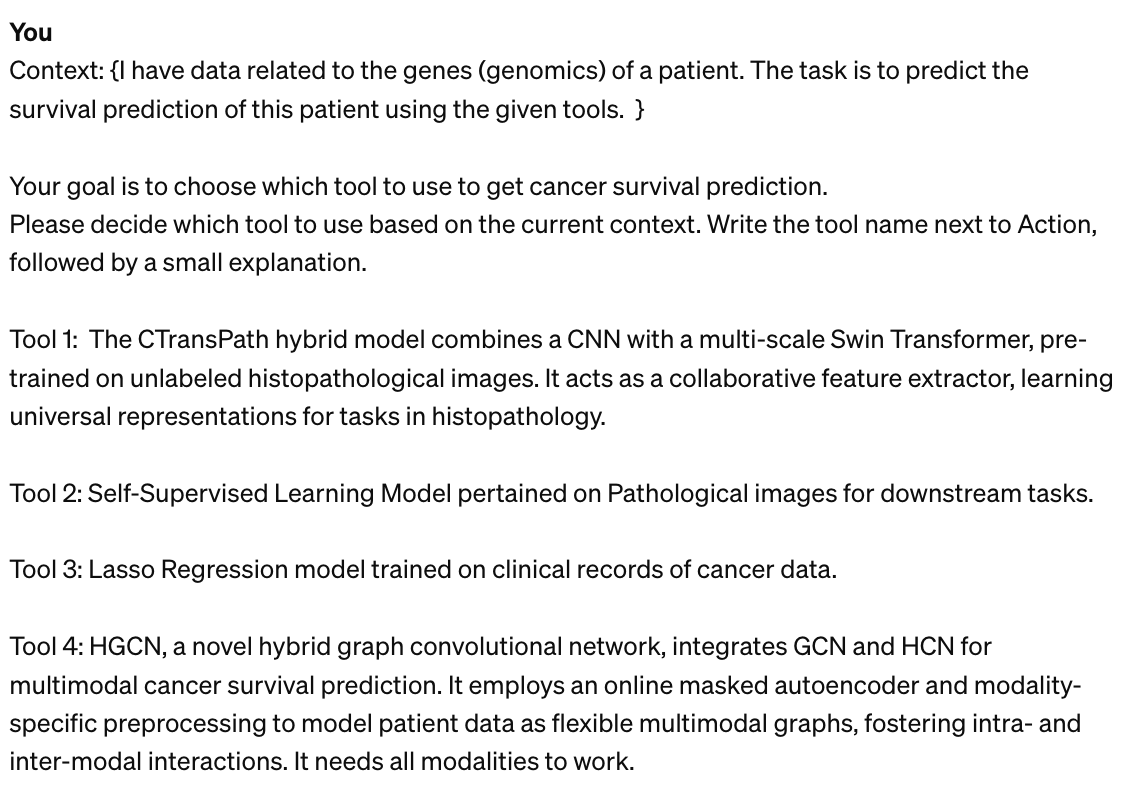}
        % \caption{Caption for image 2}
        \label{fig:sub2}
    \end{subfigure}
    \begin{subfigure}{0.49\textwidth}
        \centering
        \includegraphics[width=\linewidth]{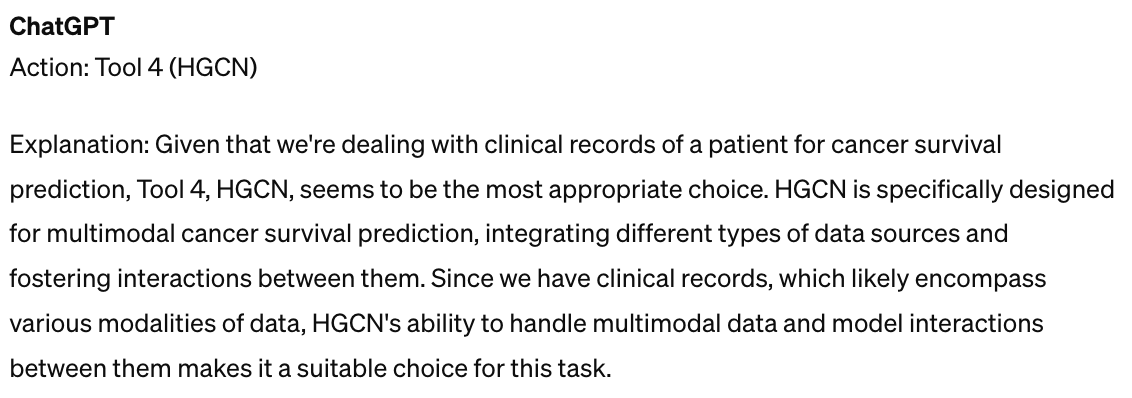}
        \caption{Example 3}
        \label{fig:sub3}
    \end{subfigure}
    \hfill
    \begin{subfigure}{0.49\textwidth}
        \centering
        \includegraphics[width=\linewidth]{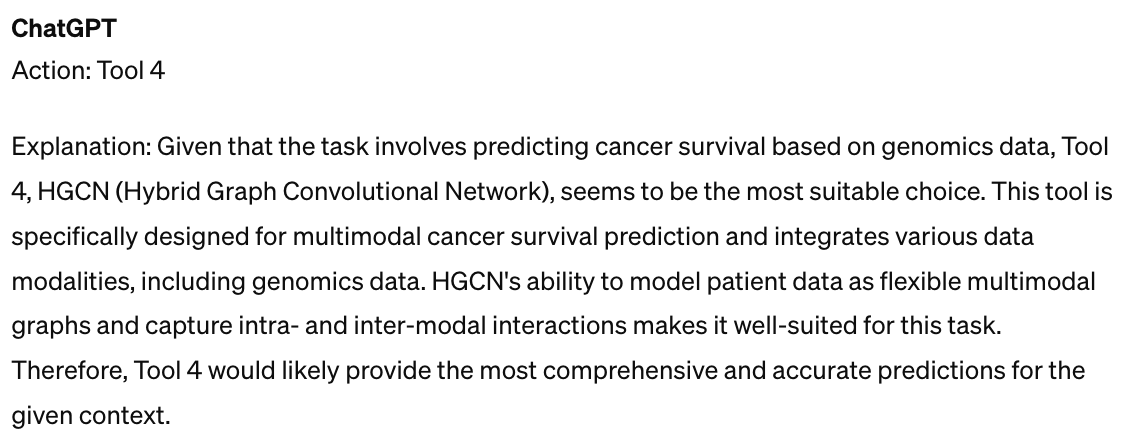}
        \caption{Example 4}
        \label{fig:sub4}
    \end{subfigure}
    % \caption{Main caption for the entire figure}
    \label{fig:main}
\end{figure}

% We observed poor performance with GPT-3.5 when asked to select a model or tool based on our available data. Example 1 illustrates that GPT-3.5 produces incorrect assumptions by hallucinating both clinical and image data, despite the context implying that only clinical data is available, leading to the selection of an inappropriate model. Tool 4 was wrongly chosen by GPT 3.5 as it requires all modalities, while Tool 2 or Tool 3 could have been chosen instead in Example 2. Example 3 and Example 4 show that when the algorithm is unable to choose a specific model, it defaults to the last tool, i.e., Tool 4. This behavior was also observed in other tested examples.
\begin{itemize}
    
\item The performance of GPT-3.5 in selecting models or tools based on available data was observed to be suboptimal in several instances. In Example 1, despite the context indicating the availability of only clinical data, GPT-3.5 produced incorrect assumptions by generating hallucinations of both clinical and image data. Consequently, this led to the selection of an inappropriate model.

\item Similarly, in Example 2, GPT-3.5 erroneously chose Tool 4, which requires all modalities, even though Tool 2 or Tool 3 could have been more suitable choices based on the available data. This suggests a failure of GPT-3.5 to accurately assess the compatibility of models with the provided data context.

\item Examples 3 and 4 further highlight GPT-3.5's tendency to default to Tool 4 when it cannot select a specific model. This behavior was consistent across multiple tested examples, indicating a limitation in the algorithm's decision-making process when confronted with uncertainty or ambiguity in the context.

\item Overall, these observations underscore the need for caution when relying on GPT-3.5 for model or tool selection in scenarios where the algorithm's understanding of the available data may be inaccurate or incomplete.

\item This observation highlights a crucial aspect: while LLMs exhibit capabilities as few-shot learners, their performance may be inadequate in specific tasks due to their lack of specific domain knowledge. 

Two key considerations arise: Firstly, the diversity of inputs across models necessitates adaptability. Secondly, while it's common to assume that one model generally outperforms another for a particular modality, there may be instances where a specific model excels, emphasizing the need for dynamic model selection. Therefore, to address the variability in model performance and the necessity for dynamic model selection, we propose a graph-guided mixture of experts-based model.
% \item This shows that although LLMs can be few-shot learners, they fail in certain domains. Hence, we explore the mixture of expert-based models and the architecture most large language models use. 
\end{itemize}

\section{Graph based - Mixture of Experts Framework}
The proposed approach for multimodal tumor survival prediction introduces a novel paradigm by employing a Graph-based Mixture of Experts (MoE) model. Unlike traditional ensemble methods where all models are trained on the same dataset and their outputs are combined through simple averaging or voting, the MoE framework offers a more sophisticated approach.

In the MoE model, each survival prediction model is treated as an "expert" specializing in a particular modality or aspect of the data. These experts are trained on subsets of the data where they can achieve optimal performance, allowing them to focus on specific features or patterns within their respective domains. This specialization enables each expert to develop a deep understanding of its assigned subset, leading to enhanced predictive capabilities.

The key innovation of the MoE framework lies in its ability to dynamically combine the outputs of these expert models based on their relevance and confidence levels for a given prediction task. Rather than relying on a fixed set of weights or voting mechanisms, the MoE dynamically selects the most suitable combination of expert predictions for each instance of the data. This adaptive aggregation mechanism ensures that the model can effectively leverage the strengths of each expert while mitigating the impact of potential weaknesses or biases.

% Moreover, the MoE framework provides a natural way to handle the inherent heterogeneity and complexity of multimodal data. By incorporating multiple expert models trained on diverse modalities such as clinical data, imaging studies, and symptom descriptions, the MoE model can capture a comprehensive range of information relevant to tumor survival prediction. This holistic approach enables the model to exploit complementary insights from different modalities, leading to more robust and accurate predictions.
Additionally, the MoE framework offers a natural means to manage the inherent heterogeneity and complexity of multimodal data. By integrating multiple expert models trained on diverse modalities like clinical data, imaging studies, and genomics, the MoE model captures a broad spectrum of relevant information for tumor survival prediction. This comprehensive approach allows the model to leverage complementary insights from various modalities, enhancing prediction robustness and accuracy.
% \textbf{The architectural elements of the proposed method:}
\section{The architectural elements of the proposed method}
\begin{figure}[htp]
    \centering
        \includegraphics[width=16cm]{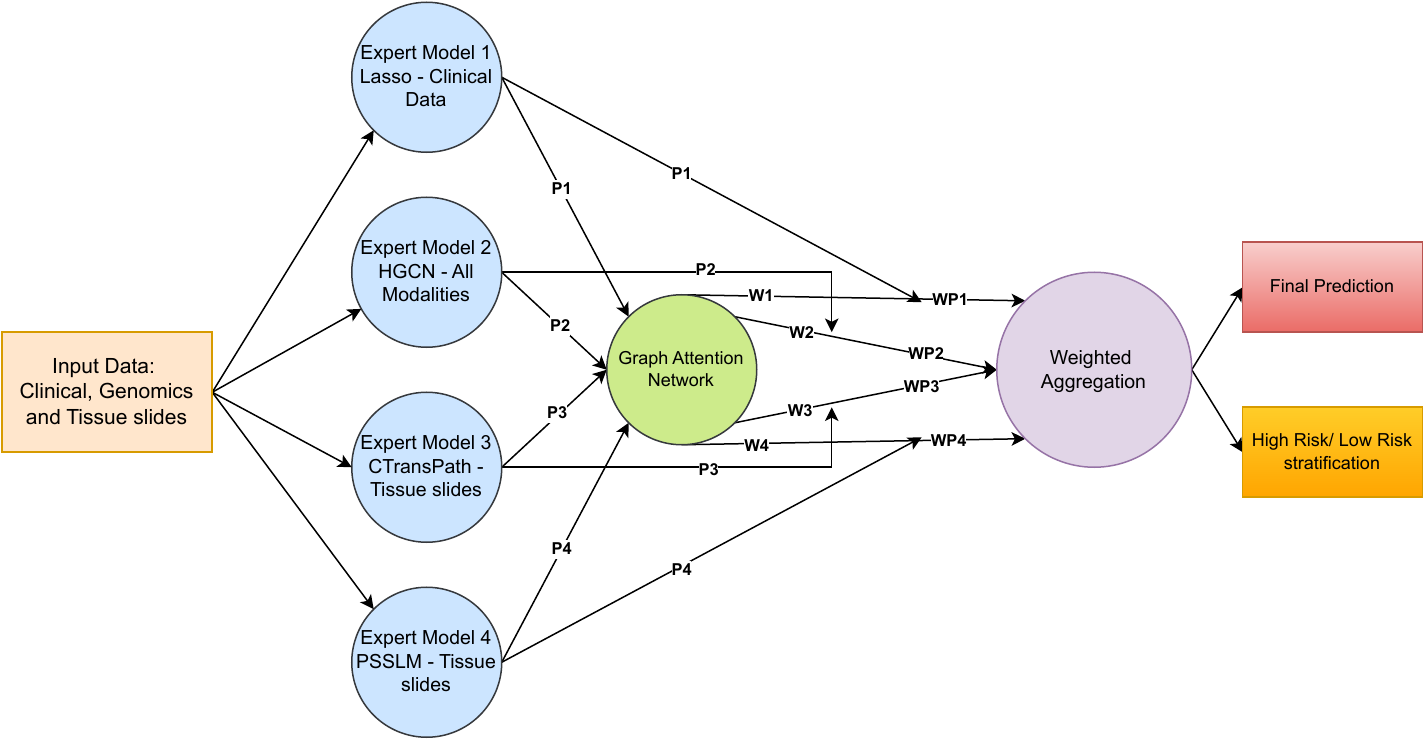}
    \captionsetup{justification=centering}
    \caption{Graph based - Mixture of Experts Framework}
    \label{fig:galaxy}
\end{figure}
% \textbf{1. Division of dataset into local subsets:} First, the predictive modeling problem is divided into subtasks based on the data a model will use. In cancer survival prediction, we work with three modalities: clinical data, genomics, and diagnostic WSIs. So, the subtasks will work on each of these modalities. 

% \textbf{2. Expert Models:} An expert is trained for each subset of the data. Typically, the experts themselves can be any model. Each expert model receives the same input pattern and makes a prediction. We have four expert models.

% \textbf{A. Lasso:} We use Lasso Regression, a popular type of regularized regression with an L1 penalty. This penalty shrinks the coefficients for input variables that do not contribute much to the prediction task. This penalty allows some coefficient values to go to zero, effectively removing input variables from the model and providing a type of automatic feature selection. The Lasso model works on a patient's clinical data.
% \textbf{1. Division of dataset into local subsets:} 
\subsection{Division of dataset into local subsets}
In the proposed approach, the predictive modeling problem is decomposed into subtasks based on the different modalities present in the data. In the context of cancer survival prediction, the data may encompass clinical information, genomic data, and diagnostic Whole Slide Images (WSIs). Each modality represents a distinct subset of the data, and subtasks are formulated to address the prediction task within each of these modalities. This division allows for focused modeling efforts tailored to the unique characteristics and challenges associated with each modality.
{
\large
\begin{equation}
x_s = \begin{bmatrix}
x_1 \\
x_2 \\
x_3 
\end{bmatrix},
\end{equation}
}

where \( x_1 \), \( x_2 \), \( x_3 \) are individual components of \( x_s \) representing clinical information, genomic data, and diagnostic Whole Slide Images (WSIs). 

% \textbf{2. Expert Models:} 
\subsection{Expert Models}

Expert models are trained to specialize in modeling each subset of the data. Each expert model will act as a domain expert for each subset of the data. These models are designed to capture the specific patterns and relationships present within their respective modalities. In the proposed framework, four expert models are utilized, each addressing a different subset of the data. We choose our expert models to cover all data modalities available in the input data. 

The expert network \( E \) is a set of \( N \) experts \( \{e_1, \ldots, e_N \} \), where each is a cancer survival prediction model and contains its own parameters in the framework. For each expert \( e_i \) (\( e_i : \mathbb{R}^D \rightarrow \mathbb{R}^D \)), it takes the token \( x_s \) as input to produce its own output \( e_i(x_s) \). 
{
\large
\begin{equation}
    m_s = E(x_s),
\end{equation}
}
% The final output of the expert network \( y_s \) is the linearly weighted combination of each expert’s output on the token by the gate’s output, formulated as Equation 2.

\subsubsection{Lasso}
% \textbf{A. Lasso:} 
One of the expert models employed is based on Lasso Regression. Lasso Regression is a variant of linear regression that incorporates an L1 penalty, which encourages sparsity in the coefficient estimates. This penalty term induces some of the coefficients to shrink to zero, effectively performing feature selection by eliminating less relevant variables. In the context of cancer survival prediction, the Lasso model operates on the patient's clinical data. By leveraging the regularization properties of Lasso Regression, this expert model can effectively identify and prioritize the most informative features from the clinical dataset, facilitating accurate predictions of survival outcomes. Additionally, the interpretability of Lasso Regression allows for a transparent understanding of the factors influencing the predictions, which can be valuable for clinical decision-making and hypothesis generation.
% \begin{figure}[htp]
%     \centering
%     \includegraphics[width=14cm]{Lasso.png}
%     % \includegraphics[width=8cm]{Lasso.pdf}
%     \captionsetup{justification=centering}
%     \caption{Lasso Regression
%  }
%     \label{fig:galaxy}
% \end{figure}
The cost function with L1 regularization is given by:
\begin{equation*}
\text{Cost function} = \frac{1}{n} \sum_{i=1}^{n} \left( (h_\theta(x)^i - y^i)^2 + \lambda \sum_{j=1}^{m} |\theta_j| \right).
\end{equation*}
where {\large $ h_{\theta}(x)^{i}$} are predicted values, {\large $ y^{i}$} are target values, $\theta_j$ is the sum of the absolute value of coefficients, and $\lambda$ is a parameter that provides a trade-off between balancing residual sum of squares and the magnitude of coefficients. 
% \textbf{B. Hybrid Graph Convolutional Network with Online Masked Autoencoder for Robust Multimodal Cancer Survival Prediction:} 
\subsubsection{Hybrid Graph Convolutional Network \citep{10061470}}
% \begin{figure}[htp]
%     \centering
%     \includegraphics[width=14cm]{HGCN.png}
%     % \includegraphics[width=8cm]{Lasso.pdf}
%     \captionsetup{justification=centering}
%     \caption{Hybrid Graph Convolutional Network
%  }
%     \label{fig:galaxy}
% \end{figure}
The Hybrid Graph Convolutional Network (HGCN) method is developed to address the challenge of robust multimodal cancer survival prediction in scenarios where patient data may be incomplete or heterogeneous across modalities. This methodology aims to enhance the representation and interaction of patient data modalities, such as clinical records, genomic features, and pathological slides, by leveraging graph-based modeling techniques. HGCN employs Graph Convolutional Networks (GCNs) to facilitate intra-modal interactions within each modality. By constructing flexible and interpretable multimodal graphs, HGCN enables effective information exchange and feature propagation within individual modalities. This intra-modal interaction mechanism allows the model to capture and incorporate relevant patterns and dependencies specific to each data modality.

In addition to intra-modal interactions, HGCN integrates a Hypergraph Convolutional Network (HCN) with a hyperedge mixing mechanism to promote inter-modal interactions. This architecture enables the model to capture high-order relationships and interactions across different modalities. By leveraging hypergraph structures, HCN facilitates the fusion of information from multiple modalities, enabling the model to extract comprehensive representations incorporating intra- and inter-modal dependencies. To further enhance the model's robustness in scenarios where certain modalities may be missing or incomplete, HGCN incorporates an Online Masked Auto-encoder. This component addresses missing modalities during model inference by capturing intrinsic dependencies between modalities and generating missing hyperedges. The Online Masked Auto-encoder enhances the model's resilience to incomplete data and improves prediction performance in real-world settings by dynamically adapting to the available modalities in a given sample.
% \textbf{B. Hybrid Graph Convolutional Network with Online Masked Autoencoder for Robust Multimodal Cancer Survival Prediction:} The Hybrid Graph Convolutional Network (HGCN) method is designed for robust multimodal cancer survival prediction. It addresses the challenge of incomplete patient data and the need for enhanced intra- and inter-modal interactions. The methodology constructs flexible and interpretable multimodal graphs, enabling effective representation and interaction of patient data modalities such as clinical records, genomic features, and pathological slides.

% HGCN leverages Graph Convolutional Networks (GCNs) to facilitate intra-modal interaction by allowing information exchange within each modality. Additionally, it integrates a Hypergraph Convolutional Network (HCN) with a hyperedge mixing mechanism to promote inter-modal interactions, capturing high-order features across modalities.

% An Online Masked Autoencoder is incorporated to address missing modalities during model inference. This component effectively captures intrinsic dependencies between modalities and generates missing hyperedges, enhancing the model's robustness in incomplete-implemented scenarios.
% \textbf{C. Transformer-based Unsupervised Contrastive Learning for Histopathological Image (CTransPath):}
\subsubsection{Transformer-based Unsupervised Contrastive Learning  \citep{WANG2022102559}}

% \begin{figure}[htp]
%     \centering
%     \includegraphics[width=14cm]{TransPATH.jpg}
%     % \includegraphics[width=8cm]{Lasso.pdf}
%     \captionsetup{justification=centering}
%     \caption{Transformer-based Unsupervised Contrastive Learning
%  }
%     \label{fig:galaxy}
% \end{figure}

The CTransPath (Transformer-based Unsupervised Contrastive Learning) method represents a pioneering approach in the realm of histopathological image analysis, aiming to extract rich and discriminative features for downstream tasks such as survival prediction. This method leverages a hybrid backbone architecture, integrating both Convolutional Neural Network (CNN) and multi-scale Swin Transformer components. By combining the strengths of CNNs in capturing local features and Transformers in modeling long-range dependencies, CTransPath effectively extracts comprehensive representations from histopathological images.

A key innovation introduced by the CTransPath method is the adoption of a semantically-relevant contrastive learning (SRCL) strategy. Unlike traditional contrastive learning approaches that align two views from an instance, SRCL focuses on enhancing instance discrimination by selecting more semantically relevant positives. This is achieved by aligning multiple positive instances that share similar visual concepts, thereby increasing the diversity of positive pairs and leading to more informative feature representations. Incorporating SRCL in CTransPath enhances both representation quality and model robustness to image variations.

The CTransPath architecture undergoes pre-training on a large-scale dataset of histopathological images using the SRCL strategy. This pre-trained model serves as a valuable asset for downstream tasks, offering two main advantages: transfer learning and direct feature extraction. Through transfer learning, the pre-trained CTransPath model can be fine-tuned on target datasets with limited labeled data, enabling efficient adaptation to specific tasks and domains. Alternatively, the pre-trained model can be used for direct feature extraction, providing rich and meaningful representations that capture salient characteristics of histopathological images.

The efficacy of the SRCL-pretrained CTransPath model has been demonstrated across various downstream tasks, showcasing state-of-the-art performance on different datasets. In the context of cancer survival prediction, we employ the CTransPath method to extract features from diagnostic slides, which are then utilized to train a regression head for predicting survival time.

\noindent
\textbf{Pathology Self-Supervised Learning Model \citep{10204656} (PSSLM)}

We employ a Self-Supervised Learning model pre-trained on diverse pathology image datasets to extract informative features from diagnostic slides for cancer survival prediction. The PSSLM model is pre-trained on a comprehensive dataset of 20,994 Whole Slide Images (WSIs) from the Cancer Genome Atlas (TCGA) and 15,672 WSIs from TULIP. These datasets encompass a wide range of cancer types and pathological variations, providing a rich source of information for feature learning.

The WSIs in these datasets are stained with Hematoxylin and Eosin (H\&E), a common staining technique used in pathology to visualize tissue structures and cell morphology. To ensure computational tractability and focus on relevant regions, patches of resolution 512x512 pixels are extracted from each WSI, resulting in a dataset containing 32.6 million patches. The pre-training process utilizes the ResNet-50 architecture, a widely used convolutional neural network known for its effectiveness in image feature extraction tasks.

By leveraging the PSSLM pre-training on such diverse and extensive pathology datasets, the model learns to capture meaningful representations of histopathological features, including tissue morphology, cellular structures, and architectural patterns. These learned representations encode rich information about the underlying biological characteristics of cancer tissues, enabling the model to discern subtle nuances and variations indicative of disease progression and prognosis.

Therefore, we utilize the learned PSSLM weights to extract features from diagnostic slides used in cancer survival prediction. The extracted features serve as input to a regression head, which is trained to predict the survival time of cancer patients.

% \textbf{3. Gating Model:}
\subsection{Gating Model}
% The gating model, also called the gating network, plays a crucial role in the Mixture of Experts (MoE) framework by interpreting the predictions generated by each expert model and determining the relative importance of these predictions for a given input. In essence, the gating model acts as a decision-maker, dynamically assigning weights to the predictions of individual expert models based on their relevance and reliability in addressing specific aspects of the input data.

In a dense MoE architecture, the gating network (graph attention network) receives the predictions produced by all expert models as input. We choose a graph-based network to enable flexibility in adding or removing experts within the framework. 
% These predictions represent different perspectives or hypotheses regarding the input data, capturing its complexity and nuances. 
The gating model then analyzes these predictions and computes corresponding weights that indicate the significance associated with each expert's output.

The key function of the gating model is to adaptively allocate attention to different expert models based on the characteristics of the input data and the task at hand. By assigning higher weights to experts whose predictions are deemed more informative or accurate for a particular input instance, the gating model enables the MoE framework to leverage the complementary strengths of diverse expert models effectively.

\begin{figure}[htp]
    \centering
    \includegraphics[width=10cm]{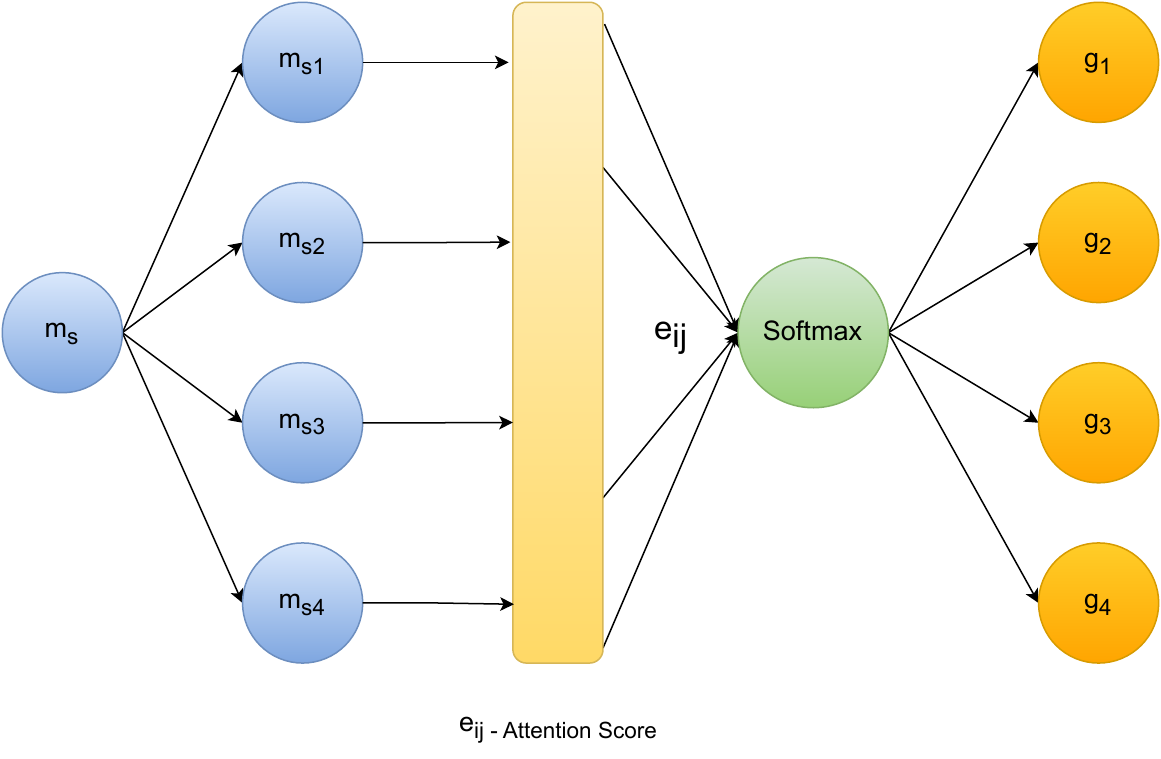}
    \captionsetup{justification=centering}
    \caption{Graph Attention Network
 }
    \label{fig:galaxy}
\end{figure}

% \textbf{3. Gating Model:} Also known as the gating network, this model helps interpret predictions made by each expert and decide which expert to trust for a given input. In the case of a dense MoE, the gating network takes in the predictions of all the models and assigns weights for each prediction. 

The dense gate network \( G \) takes the expert prediction as input \( y_s \) (\( y_s \in \mathbb{R}^D \)) and produces the probability of it with respect to \( N \) experts. Equation 1 formulates the gate function, where \( y_s \) (\( y_s \in \mathbb{R}^{D \times N} \)) are predictions of \( N \) experts, \( A_s \) is the Attention Network and the outputs are normalized via softmax distribution. 

{\large
\begin{equation}
    g(x_s) = \text{softmax}(A_s(m_s)),
\end{equation}}

% \textbf{4. Pooling Method: }
\subsection{Pooling Method}
Following the gating model's determination of the weights for each expert's prediction, a pooling method is employed to combine these weighted predictions into a single aggregated output. The pooling method serves as the final step in the MoE framework, synthesizing the diverse insights provided by individual experts into a cohesive and comprehensive prediction.

One common approach to pooling in the MoE framework is weighted sum or weighted aggregation. In this method, the gating model's weights are multiplied by the corresponding expert model outputs and summed to produce the final aggregated prediction. Weighted sum pooling ensures that the contributions of each expert model are appropriately weighted based on their respective strengths and relevance to the input data. By incorporating the gating mechanism and weighted aggregation, the graph based - MoE framework harnesses the collective intelligence of multiple expert models, resulting in enhanced predictive performance and robustness across a wide range of tasks and datasets. 

The final output is the linearly weighted combination of each expert’s output on the token by the gate’s output.
% \[
% y_s = \sum_{i=1}^{N} g(x_s) \cdot e_i(x_s)
% \]
% \[
% \bigg( y_s = \sum_{i=1}^{N} g(x_s) \cdot e_i(x_s) \bigg)
% \]
{\large
\begin{equation}
   % \[
\displaystyle y_s = \sum_{i=1}^{N} g(x_s) \cdot e_i(x_s).
% \] 
\end{equation}}

\chapter{EXPERIMENTS}
% \textbf{A. Dataset:}
\section{Dataset}
The Cancer Genome Atlas (TCGA) is a vital dataset that has provided molecular information on more than 20,000 primary cancers and matched normal samples, covering 33 different types of cancer. TCGA data has significantly improved the ability to diagnose, treat, and prevent cancer, and has also changed the way in understanding and treating cancer in clinics. We selected the TCGA-LUAD dataset, which covers lung adenocarcinoma (LUAD), as it is one of the most common cancer types. LUAD prognosis is still a significant clinical challenge, as it varies remarkably among individual patients.
% The Cancer Genome Atlas (TCGA) is a benchmark dataset that molecularly characterized over 20,000 primary cancers and matched normal samples spanning 33 cancer types. TCGA data has improved the ability to diagnose, treat, and prevent cancer and has also changed how cancer is understood and treated in the clinic. We chose the TCGA-LUAD (lung adenocarcinoma (LUAD)) cancer dataset from the Genomic Data Commons (GDC) Data Portal as it is one of the most common cancer types. The prognosis of LUAD is still a major clinical challenge, which is remarkably different among individual patients with LUAD. 
The dataset contains 585 cases of data, and after merging clinical data, 470 cases in the TCGA-LUAD database with genomics, Imaging Slides, and clinical data were used for analysis.

\begin{table}[htbp]
    \centering
    \caption{Details of TCGA-LUAD Dataset}
    \begin{tabular}{|c|c|}
        \hline
        \rowcolor{gray!30}
        \multicolumn{2}{|c|}{Clinical Records of TCGA-LUAD (N=470)} \\
        \hline
        \textbf{Attribute} & \textbf{Description} \\
        \hline
        Race & Race of patients \\
        Age at Index & Age of patients at diagnosis \\
        Gender & Gender of patients \\
        Pathological stage & Cancer stage based on pathology reports \\
        Morphology & Cellular structure and characteristics \\
        Radiation Therapy & Whether patients underwent radiation therapy \\
        Pharmaceutical Therapy & Pharmaceutical treatments received \\
        \hline
    \end{tabular}
    \label{tab:clinical_records}
\end{table}

\begin{table}[h]
    \centering
    \caption{Diagnostic Slide Transformation}
    \begin{tabular}{|c|c|}
        \hline
        \rowcolor{gray!30}
        \multicolumn{2}{|c|}{Diagnostic Slide Transformation} \\
        \hline
        \textbf{Transformation} & \textbf{Details} \\
        \hline
        Slide Transformation & .svs to .png \\
        Mean & $(0.485, 0.456, 0.406)$ \\
        Standard Deviation & $(0.229, 0.224, 0.225)$ \\
        Resize & $224 \times 224$ pixels \\
        \hline
    \end{tabular}
    \label{tab:slide_transformation}
\end{table}

\section{Baselines and Metrics}

We use the four expert models, i.e., Lasso, HGCN, CTransPath, and PSSLM, as baselines, including vanilla ensemble. 

\textbf{Metrics:} The performance of the models is assessed using two metrics:

\textbf{1.} \textbf{The concordance index (c-index)}, which evaluates the extent to which the model's predicted values correctly rank the observed survival times. A higher c-index indicates better predictive performance, with values ranging from \textbf{0.5 }(random prediction) to $\textbf{1}$ (perfect prediction).
\begin{equation*}
c = \frac{\sum_{i \in U}\Bigl\{ \sum_{T_j > T_i} 1_{f_j > f_i}\Bigl\} }{\sum_{i \in U} \Bigl\{\sum_{T_j > T_i} 1\Bigl\}},
\end{equation*}

where $U$ is a set of uncensored data; $T_i$ is the observed survival time of sample $i$; $f_i$ is the predicted survival time of sample $i$ and $1_{a>b} = 1$ if $a>b$ and $0$ otherwise.
% \begin{figure}[h]
%     \centering
%     \includegraphics[width=6cm]{cindex.png}
%     % \includegraphics[width=8cm]{Lasso.pdf}
%     \captionsetup{justification=centering}
%     \caption{c-index
%  }
%     \label{fig:galaxy}
% \end{figure}

\textbf{2.} \textbf{Kaplan-Meier (KM) curves}, which visualize the probability of survival for patients in different risk groups over a specific time period. Additionally, we employ the logrank test to assess the statistical significance of the differences between low and high-risk groups. A \textbf{p-value} of less than $\textbf{0.05}$ indicates a statistically significant separation between the groups.
\begin{align*}
\tau_1 &< \tau_2 < \cdots < \tau_K \text{ are distinct failure times} \\
Y_i(\tau_j) & \text{ are number of persons in group } i \text{ at risk at } \tau_j \\
Y(\tau_j) & = Y_0(\tau_j) + Y_1(\tau_j), \text{ represents the total number of subjects at risk at } \tau_j \\
d_{ij} & \text{ is the number of failures in group } i \text{ at } \tau_j \\
d_j & = d_{0j} + d_{1j} \text{ are total number of failures at } \tau_j
\end{align*}

\begin{itemize}
    % \item \tau_1 &< \tau_2 < \cdots < \tau_K \text{ are distinct failure times} 
    % \item Y_i(\tau_j) & = \text{number of persons in group } i \text{ at risk at } \tau_j 
    % \item Y(\tau_j) & = Y_0(\tau_j) + Y_1(\tau_j), \text{ the total number of subjects at risk at } \tau_j 
    % \item d_{ij} & = \text{number of failures in group } i \text{ at } \tau_j 
    % \item d_j & = d_{0j} + d_{1j} \text{ total number of failures at } \tau_j
    \item $O_j$ is the observed number of failures
    \item $E_j$ is the expected number of failures, where $E_j = d_j Y_1(\tau_j) / Y(\tau_j)$
    \item $V_j = \frac{Y_0(\tau_j) Y_1(\tau_j) d_j (Y(\tau_j) - d_j)}{Y(\tau_j)^2 (Y(\tau_j) - 1)}$ is the variance of the observed number of failures
    \item $Z$ represents the p-value. 
\end{itemize}
\[
Z = \frac{\sum_{j=1}^{k} (O_j - E_j)}{\sqrt{\sum_{j=1}^{k} V_j}}.
\]
\section{Implementation}

In this research, we conducted a 5-fold evaluation procedure to assess the performance of survival prediction methods, considering the variability of predictions. For evaluation purposes, 20\% of the data was randomly sampled as a test set, and the average scores were reported. The models were implemented using PyTorch. 

\textbf{1. }For the Lasso model, the Sklearn Lasso implementation was utilized with hyperparameters set to $\textbf{alpha=10}$ and $\textbf{max\_iteration=2000}$.

\textbf{2. } The regression head architecture for the CTransPath and PSSLM models consists of a multi-layer perceptron (MLP) with three fully connected layers: fc1 with 128 neurons, fc2 with 64 neurons, and fc3 for the final regression output. Each layer is followed by a Rectified Linear Unit (ReLU) activation function to introduce non-linearity. During the training phase, the Mean Squared Error (MSE) loss function is utilized to quantify the difference between predicted and actual regression values. The Adam optimizer is employed to update the model parameters, using a learning rate of $\textbf{0.001}$. A batch size of $\textbf{16}$ is employed, and data shuffling is enabled to facilitate model convergence and mitigate the risk of overfitting.

% The regression head architecture for the CTransPath and PSSLM models comprises three fully connected layers $\textbf{(fc1, fc2, fc3)}$ with Rectified Linear Unit (ReLU) activation functions, followed by a linear output layer. Initially, the input data undergoes flattening to convert the multi-dimensional feature vectors of tissue slides into a 1D tensor. This flattened input is then passed through the first fully connected layer $\textbf{(fc1)}$ consisting of 128 neurons, where a ReLU activation function introduces non-linearity. Subsequently, the output of fc1 is fed into the second fully connected layer $\textbf{(fc2)}$ with 64 neurons, followed by another ReLU activation function. Finally, the output of $\textbf{(fc2)}$ is processed through the last fully connected layer $\textbf{(fc3)}$, which generates the final regression output. During the training phase, the Mean Squared Error (MSE) loss function is utilized to quantify the difference between predicted and actual regression values. The Adam optimizer is employed to update the model parameters, using a learning rate of $\textbf{0.001}$. A batch size of $\textbf{16}$ is employed, and data shuffling is enabled to facilitate model convergence and mitigate the risk of overfitting.

\textbf{3. }The Graph-based - Mixture of Experts (MoE) architecture incorporates a Gating Function and MoE Function. 

% This GatingFunction consists of two fully connected layers $\textbf{(fc1 and fc2)}$ with Rectified Linear Unit (ReLU) activation functions. 
The Gating Function operates on the concatenated predictions from all expert models. The concatenated predictions undergo processing through two fully connected layers: fc1 followed by fc2. fc1 employs a Rectified Linear Unit (ReLU) activation function, while fc2 applies a softmax activation function along the expert dimension. This softmax operation generates gating coefficients, ensuring their sum equals one and indicating the relative contribution of each expert's prediction.

% The GatingFunction takes the concatenation of predictions from all expert models as input. Upon passing through $\textbf{fc1}$, the output is activated by ReLU, ensuring non-linearity in the transformation. Subsequently, the output of $\textbf{fc1}$ is fed into $\textbf{fc2}$, followed by a softmax activation function along the expert dimension. This softmax operation yields gating coefficients, ensuring their sum equals one and representing the relative contribution of each expert's prediction.
The MoE function encapsulates the model, taking the gating function as input during initialization. Here, gating coefficients are computed using the gating function and applied to expert predictions to generate the final output. In training, the Adam optimizer updates model parameters with a learning rate of $\textbf{0.1}$ and a batch size of $\textbf{16}$ with data shuffling is employed to enhance convergence and mitigate overfitting.
\\
In our experiments, we aim to address the following questions:
\begin{itemize}
    \item \textbf{RQ.1:} How does the Graph-Guided Mixture of Experts (MoE) framework improve the accuracy of survival prediction?
    \item \textbf{RQ.2:} What is the relative importance of each expert model in enhancing the predictive capability of the MoE framework?
\end{itemize}
% In the MoE Function, the model is encapsulated, taking the gating function as input during initialization. During the forward pass, this class computes gating coefficients utilizing the gating function and applies them to the predictions from expert models to generate the final prediction. This process involves a weighted combination of expert predictions based on the gating coefficients, resulting in a unified output tensor. The gating coefficients dynamically adjust the influence of each expert's prediction, allowing the model to adapt effectively to varying input conditions. During the training phase, the Adam optimizer updates model parameters with a learning rate $\textbf{0.1}$. Additionally, a batch size of $\textbf{16}$ and shuffling of data are employed to enhance model convergence and mitigate overfitting risks.

% \textbf{C. Baselines and Metrics:}

% The performance of the models is assessed using two metrics. Firstly, the concordance index (c-index) which measures how well the model predicts the actual survival rate. The c-index ranges from 0.5 (random prediction) to 1 (perfect prediction), and higher values indicate better performance. Secondly, Kaplan-Meier curves are used to visualize the survival probability of patients belonging to different risk groups for a certain period. To determine if there is a significant difference in survival rates between low and high-risk groups, the log-rank statistical significance test is applied with a threshold of p-value less than 0.05.
\chapter{RESULTS and ANALYSIS}

% \textbf{1. ChatGPT (3.5)}

% \begin{table}[htbp]
% \begin{tabularx}{\textwidth}{|l|X|X|}
% \hline
% Type & Method & LUAD \\
% \hline
% \multirow{4}{*}{Individual Models} & Lasso & 0.650+0.017 \\
% & HGCN & 0.663+0.020 \\
% & CTransPath & 0.611+0.085 \\
% & SSL & 0.642+0.027 \\
% \hline
% Ensemble & Naive & 0.665+0.020 \\
% \hline
% \begin{tabular}[c]{@{}l@{}}Mixture of \\ Experts\end{tabular} & Dense & 0.703+0.022 \\
% \hline
% \end{tabularx}
% \end{table}
% \textbf{2. Mixture of Experts}
% \section{Mixture of Experts}
\section{RQ.1 Performance Comparison}
\begin{table}[htbp]
\centering
\caption{The results of (CI $\uparrow$) of Individual Models, comparison with Ensemble and MoE based model.}\label{tab1}
\begin{tabularx}{\linewidth}{|>{\centering\arraybackslash}X|>{\centering\arraybackslash}X|>{\centering\arraybackslash}X|}
\hline
\rowcolor{gray!30}
\textbf{Type} & \textbf{Method} & \textbf{LUAD} \\ \hline
\multirow{4}{*}{\textbf{Individual Models}} & Lasso & 0.650$\pm$0.017 \\ \cline{2-3} 
& HGCN & 0.633$\pm$0.020 \\ \cline{2-3} 
& CTransPath & 0.611$\pm$0.085 \\ \cline{2-3} 
& PSSLM & 0.642$\pm$0.027 \\ \hline
\textbf{Ensemble} & Naive & 0.665$\pm$0.020 \\ \hline
\begin{tabular}[c]{@{}c@{}}\textbf{Mixture of Experts}\end{tabular} & Dense & \textbf{0.703$\pm$0.022} \\ \hline
\end{tabularx}
\end{table}
% \begin{table}[htbp]
% \caption{The results of (CI $\uparrow$) of Individual Models, comparison with Ensemble and MoE based model.}\label{tab1}
% \begin{tabularx}{\linewidth}{|X|X|X|}
% \hline
% \rowcolor{gray!30}
% \textbf{Type }                                                         & \textbf{Method}     & \textbf{LUAD }       \\ \hline
% \multirow{4}{*}{\textbf{Individual Models}}                            & Lasso      & 0.650$\pm$0.017 \\ \cline{2-3} 
%                                                               & HGCN       & 0.663$\pm$0.020 \\ \cline{2-3} 
%                                                               & CTransPath & 0.611$\pm$0.085 \\ \cline{2-3} 
%                                                               & SSL        & 0.642$\pm$0.027 \\ \hline
% \textbf{Ensemble}                                                      & Naive      & 0.665$\pm$0.020 \\ \hline
% \begin{tabular}[c]{@{}l@{}}\textbf{Mixture of } \textbf{Experts}\end{tabular} & Dense      & \textbf{0.703$\pm$0.022} \\ \hline
% \end{tabularx}
% \end{table}

\begin{itemize}
  
\item As depicted in Table~\ref{tab1}, our analysis begins by assessing the ranking ability, quantified by the Concordance Index (CI), across various methods. Initially, it is evident that individual modal performances tend to exhibit lower CI values compared to multimodal approaches. This disparity underscores the advantage of leveraging diverse and complementary information from multiple modalities in the context of survival prediction tasks.

\item Among the Individual Models assessed, Lasso and PSSLM stand out as the most effective methods. This outcome highlights the ability of Lasso and PSSLM to leverage information from clinical records and image slides, respectively, thereby augmenting their predictive ability.
% \item Among the Individual Models evaluated, the Hybrid Graph Convolutional Network (HGCN) emerges as the top-performing method. This result underscores the efficacy of HGCN in harnessing information from all available modalities, thereby enhancing its predictive capabilities.

\item Subsequently, employing a naive ensemble technique yields a modest improvement in CI compared to the Individual Models. Despite the marginal enhancement, it reinforces the utility of ensemble methods in amalgamating predictions from diverse sources to achieve more robust performance.

\item However, our proposed method, which strategically amalgamates individual modal and multimodal approaches, attains the pinnacle of performance. This integration allows for the exploitation of singular and combined sources of information, culminating in a mean CI of $\textbf{0.703}$. Importantly, this surpasses the performance of both the Individual Models and the Ensemble, highlighting the effectiveness of our approach in leveraging the complementary strengths of different survival prediction models.

\item In the analysis of the TCGA-LUAD (The Cancer Genome Atlas - Lung Adenocarcinoma) dataset, we employed a risk stratification approach based on the predictions generated by our model. The goal is to categorize patients into high-risk and low-risk groups according to their predicted survival outcomes. This stratification allows for a clearer understanding of patient prognosis and can inform treatment decisions. We evaluate several models, including individual modal models and an ensemble model, for predicting patient survival in the TCGA-LUAD dataset. Each model generated its own set of predictions, allowing us to compare their performance in stratifying the patients into high-risk and low-risk groups.

\item Fig.~\ref{fig:main1}-~\ref{fig:main3} illustrates the Kaplan-Meier (KM) survival curves for the high-risk and low-risk groups identified by different models. These curves demonstrate the estimated probability of survival over time for each group. Among the individual models are lasso and PSSLM, which exhibit higher performance compared to others. However, when considering the ensemble model, which combines the predictions of multiple individual models, we observed improved performance in patient stratification. Interestingly, the ensemble model demonstrated the ability to stratify patients more effectively than any of the individual models alone. This suggests that leveraging the diversity of predictions from multiple models can enhance the accuracy of risk stratification and provide more reliable insights into patient prognosis.

\item The Mixture of the Expert-based model, which we utilized in our analysis, exhibited state-of-the-art (SOTA) performance in predicting patient survival within the LUAD dataset.

\item Comparing the KM curves of the high-risk and low-risk groups reveals the significance of our method's performance. A key indicator of the effectiveness of our model is the p-value associated with the differences between these curves. A low p-value suggests a significant difference in survival outcomes between the two groups. In our analysis, the p-values obtained from our method were consistently lower than those from the other methods, generally less than 0.01.

\item This indicates that our model successfully stratifies patients into distinct risk groups with significantly different survival outcomes. Such findings highlight the effectiveness of our approach for patient stratification in the context of lung adenocarcinoma, providing valuable insights for clinical decision-making and personalized treatment strategies.

\begin{figure}
    \centering
    \begin{subfigure}{0.49\textwidth}
        \includegraphics[width=\textwidth]{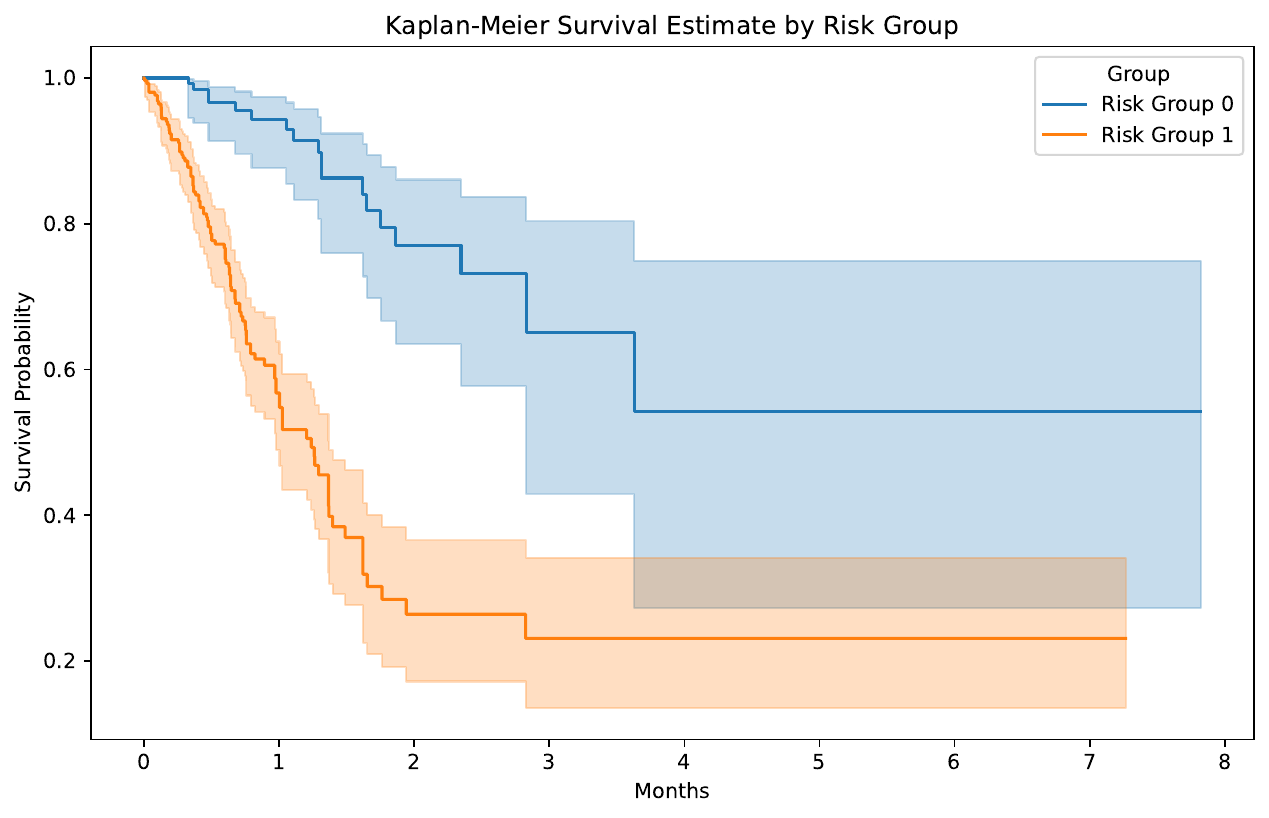}
        \caption{p-value = $4.97 \times 10^{-13}$}
        \label{fig:sub11}
    \end{subfigure}
    \hfill
    \begin{subfigure}{0.49\textwidth}
        \includegraphics[width=\textwidth]{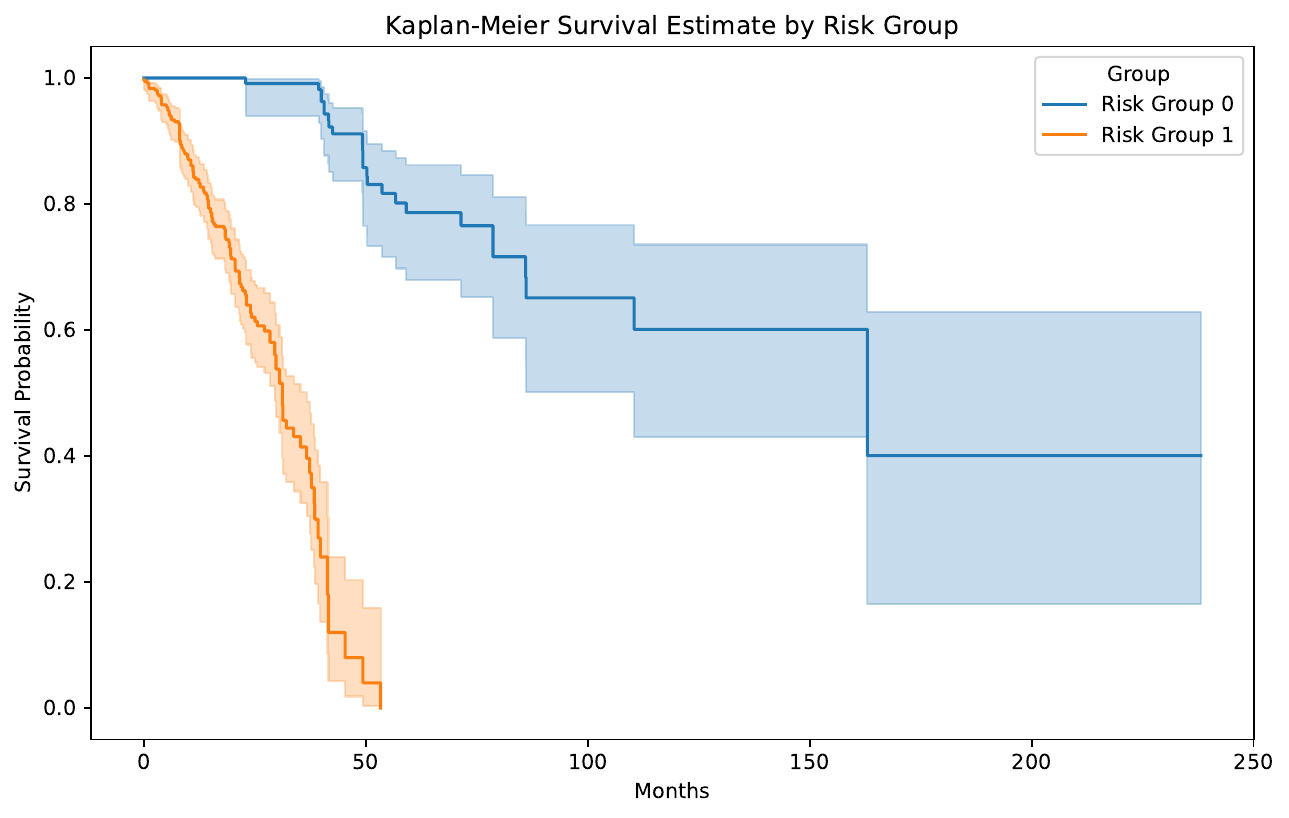}
        \caption{p-value = $8.86 \times 10^{-29}$}
        \label{fig:sub21}
    \end{subfigure}
    \caption{a) HGCN, b) Lasso}
    \label{fig:main1}
\end{figure}

\begin{figure}
    \centering
    \begin{subfigure}{0.49\textwidth}
        \includegraphics[width=\textwidth]{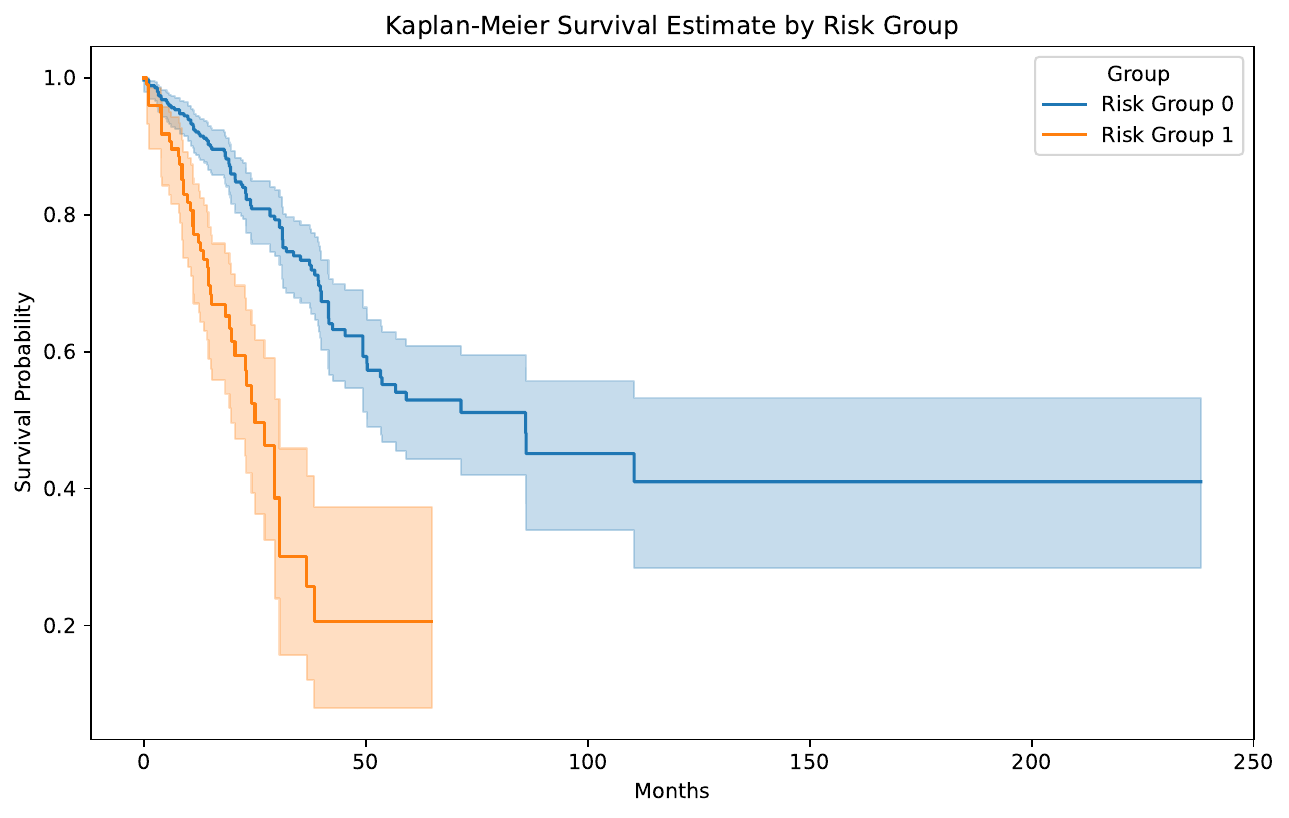}
        \caption{p-value = $2.41\times10^{-11}$}
        \label{fig:sub12}
    \end{subfigure}
    \hfill
    \begin{subfigure}{0.49\textwidth}
        \includegraphics[width=\textwidth]{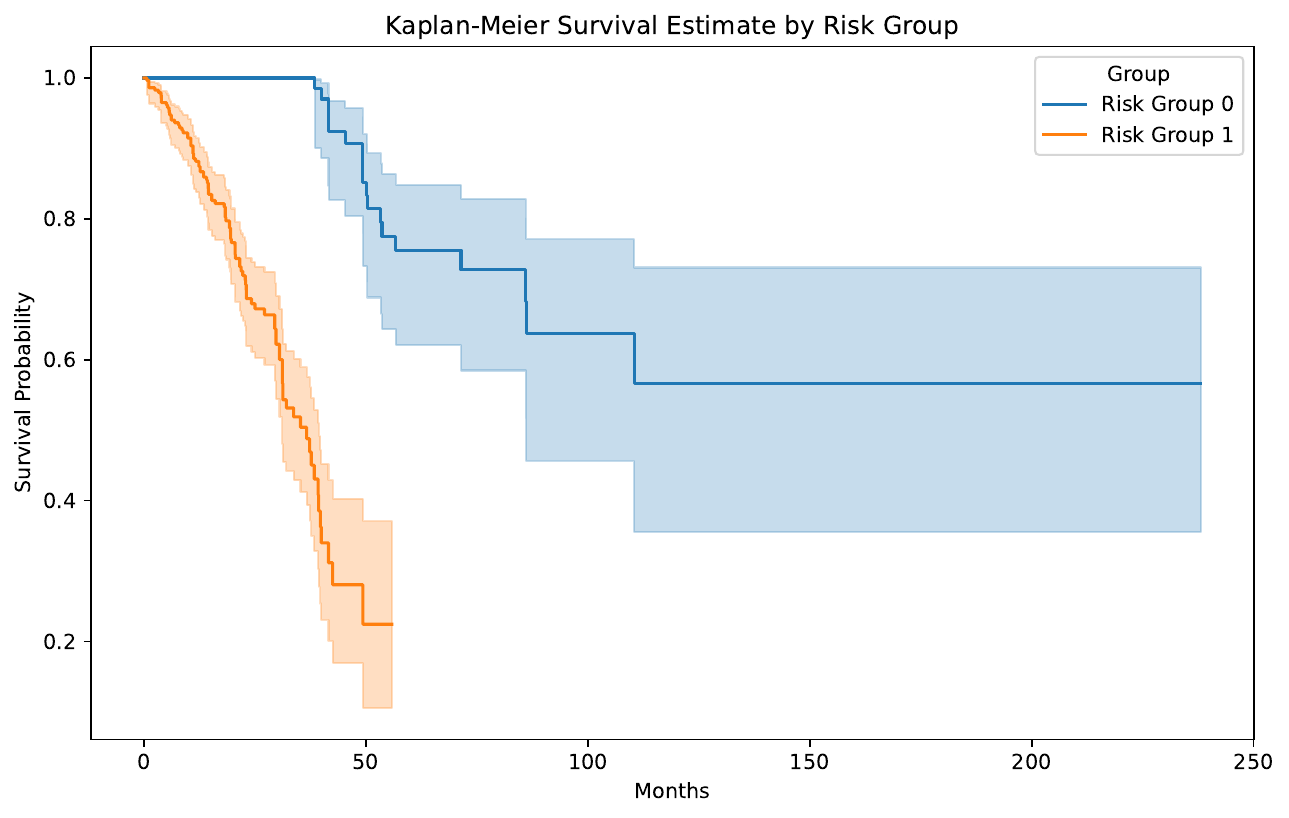}
        \caption{p-value = $1.05\times10^{-14}$}
        \label{fig:sub22}
    \end{subfigure}
    \caption{a) CTransPath, b) PSSLM}
    \label{fig:main2}
\end{figure}
\begin{figure}
    \centering
    \begin{subfigure}{0.49\textwidth}
        \includegraphics[width=\textwidth]{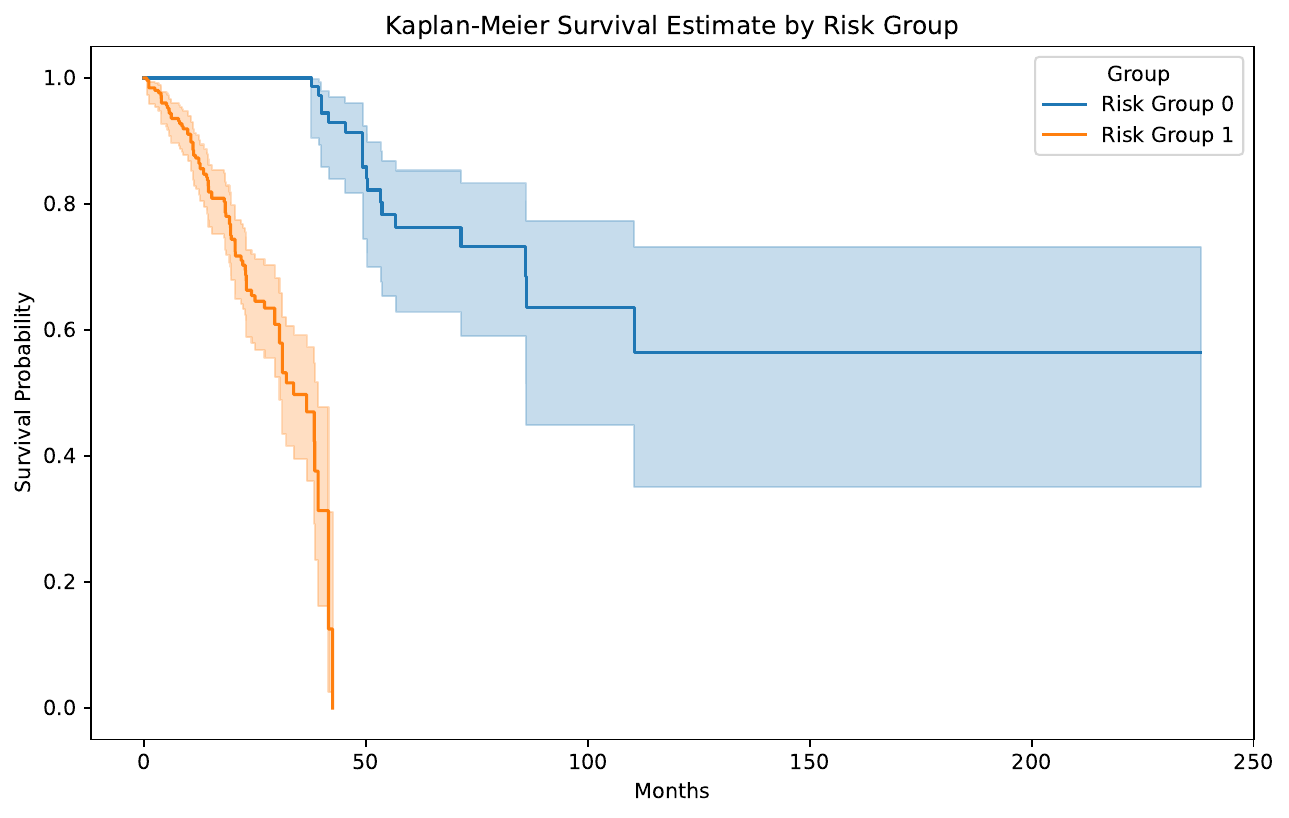}
        \caption{p-value = $2.31\times10^{-16}$}
        \label{fig:sub13}
    \end{subfigure}
    \hfill
    \begin{subfigure}{0.49\textwidth}
        \includegraphics[width=\textwidth]{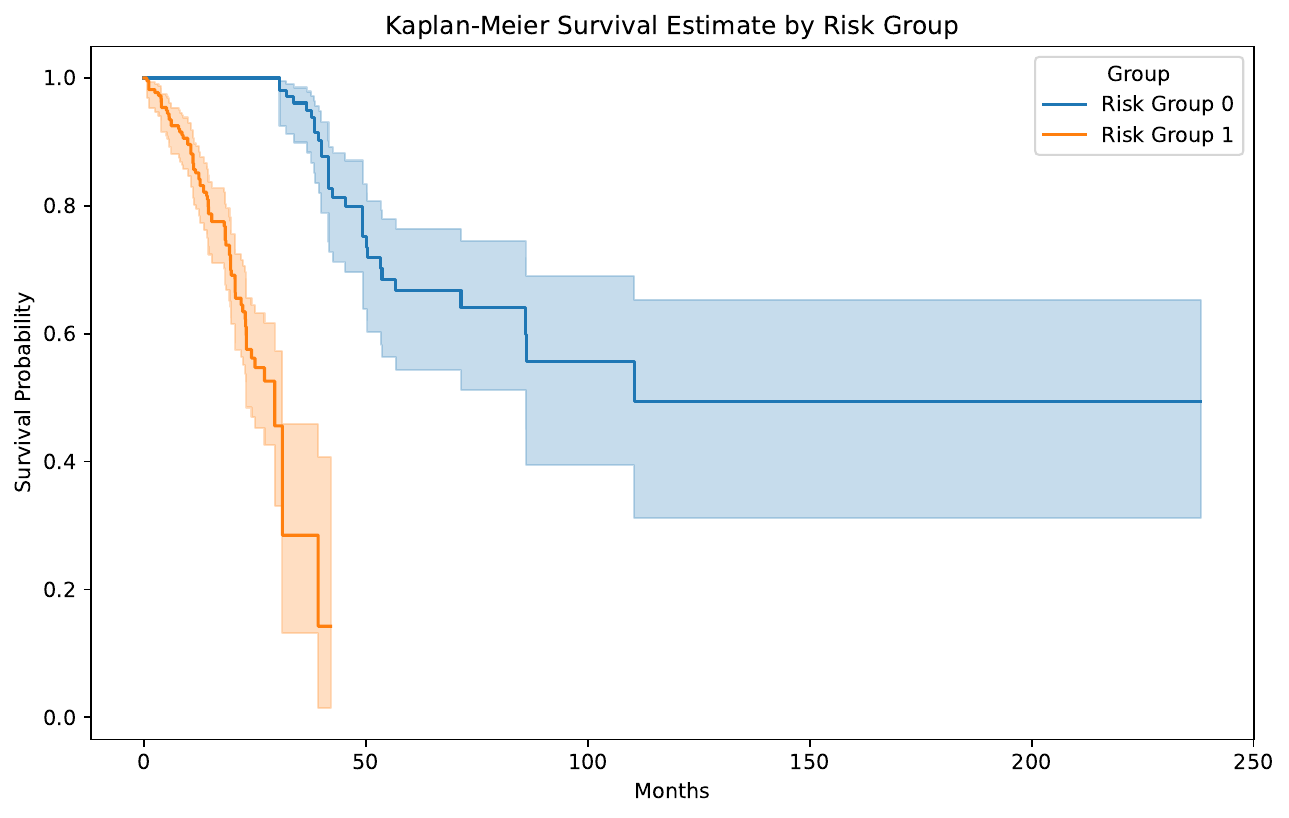}
        \caption{p-value = $5.10\times10^{-19} $}
        \label{fig:sub23}
    \end{subfigure}
    \caption{a) Ensemble, b) Mixture of Experts(MOE)}
    \label{fig:main3}
\end{figure}
\end{itemize}

\newpage

\section{RQ.2: Relative Significance of Each Expert Model}
\begin{figure}
    \centering
    \begin{subfigure}{0.49\textwidth}
        \includegraphics[width=\textwidth]{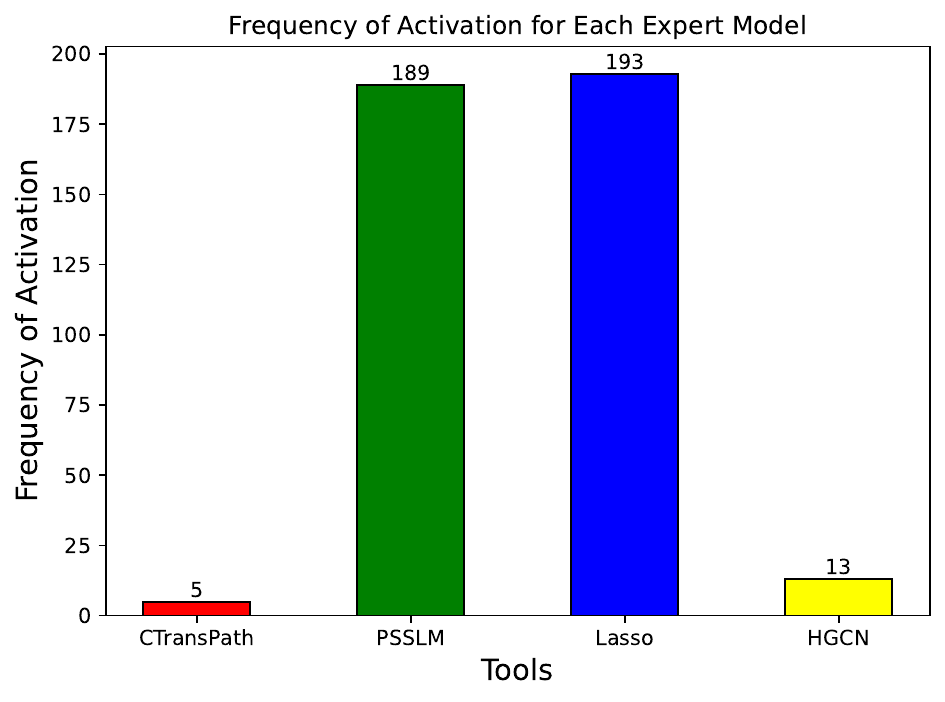}
        \caption{Frequency of Activation}
        \label{fig:sub1}
    \end{subfigure}
    \hfill
    \begin{subfigure}{0.49\textwidth}
        \includegraphics[width=\textwidth]{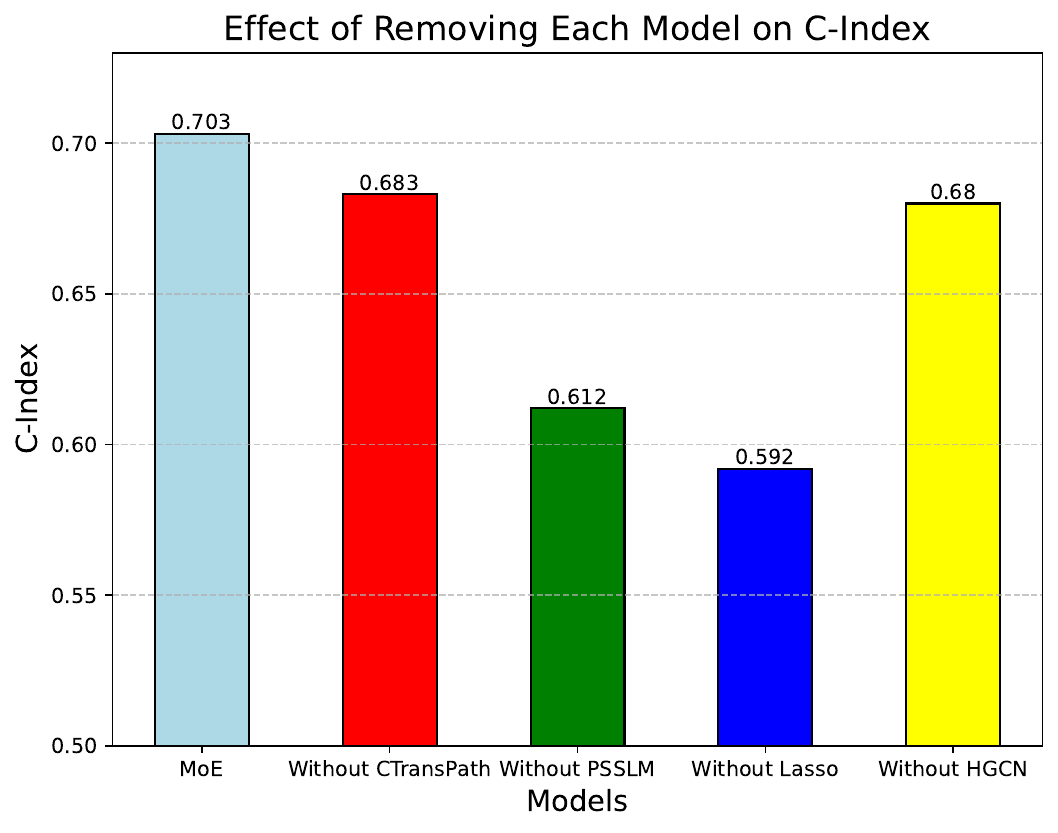}
        \caption{Effect of Removing Each Model}
        \label{fig:sub2}
    \end{subfigure}
    \caption{Significance of Each Expert Mode}
    \label{fig:main}
\end{figure}

\begin{itemize}
\item Figure~\ref{fig:sub1} illustrates the frequency of activation for each expert model within our framework. Notably, Lasso and PSSLM exhibit the highest frequency of activation compared to the other two models. This observation can be attributed to the robustness and effectiveness of PSSLM and Lasso in survival prediction tasks, which positions them as preferred choices within our framework. The frequency of activation reflects the adaptability of our framework, as it dynamically selects the most suitable expert model based on the patient's input data. Consequently, the increased frequency of activation for superior models underscores our framework's ability to prioritize and leverage the strengths of high-performing models for improved predictive outcomes.
\end{itemize}

% From Figure ~\ref{fig:main}, we can see that both are complementing each other, underscoring the fact that the framework dynamically selects the high performing model for a particular input data and how a removal of high and low performing models when removed has effect within the framework. 
% \begin{table}[htbp]
% \centering
% \caption{Effect of Removing Each Model on C-Index}
% \label{tab:effect-of-removing-model}
% \begin{tabularx}{\linewidth}{|>{\centering\arraybackslash}X|>{\centering\arraybackslash}X|}
% \hline
% \rowcolor{gray!30}
% \textbf{Model Removed} & \textbf{C-Index} \\
% \hline
% Without CTransPath & 0.683 \\
% Without HGCN & 0.680 \\
% Without Lasso & 0.592 \\
% Without SSL & 0.612 \\
% \hline
% \end{tabularx}
% \end{table}

From Figure~\ref{fig:sub2}, we can make the following analysis:
\begin{itemize}
     \item The predictive performance of a framework that utilizes various models can be significantly impacted by the removal of specific models. In this context, the C-Index decreases notably if the Lasso model is removed, which highlights its crucial role in enhancing the accuracy of survival prediction. 

\item Similarly, the PSSLM model also shows a considerable impact on the predictive performance, as removing it leads to a noticeable decrease in the C-Index. This suggests that the PSSLM model effectively harnesses information from image slides, contributing significantly to the overall predictive prowess of the framework.

\item Furthermore, removing the HGCN model also results in a notable decrease in the C-Index, implying its importance in the predictive capability of the framework. On the other hand, the removal of the CTransPath model has a relatively minor effect on the C-Index compared to the other models. While still contributing to the overall predictive performance, its removal does not result in as significant a decrease in the C-Index as observed with the other models.
\end{itemize}

\noindent
In conclusion, the analysis of Figure~\ref{fig:main} reveals a complementary relationship between the two, highlighting the framework's ability to select the most suitable model based on input data dynamically. Additionally, the impact of removing both high and low-performing models underscores the framework's adaptability and effectiveness. Due to the inductive nature of our gating model, our framework is also flexible, capable of adding more experts or removing experts. 
\chapter{CONCLUSION and LIMITATIONS}
\section{CONCLUSION}
In conclusion, this thesis has addressed the challenges and limitations faced by the existing survival prediction models in multi-modal tumor survival prediction. Through the development of a novel approach utilizing a graph-guided Mixture of Experts (MoE) framework, we have demonstrated the potential to enhance predictive power and accuracy for survival prediction tasks. By leveraging the strengths of existing models and guiding them through a MoE framework, the proposed approach offers a promising solution to the integration of multi-modal data and the effective management of model callings. The experiments conducted on the TCGA-LUAD dataset have shown improved performance over individual modal and vanilla ensemble models, highlighting the effectiveness of our approach in addressing the challenges posed by complex real-world tasks.
\newpage
\section{LIMITATIONS}
% LIMITATIONS: 
Although our research has yielded promising results, it is important to acknowledge that the analysis and experiments were conducted solely on the TCGA LUAD dataset. As such, we cannot guarantee the generalizability of our approach across diverse datasets and application domains.

Furthermore, our study is limited in terms of interpretability. While our approach allows for survival prediction by tracing back to specific data modalities, it does not currently offer the ability to pinpoint the exact biological and clinical factors or features within each modality that contribute to the predictions. This lack of granularity in feature attribution within individual modalities restricts our ability to provide detailed insights into the underlying mechanisms driving the model's predictions.

Another limitation we encountered concerns the computational cost associated with adding more experts to the model. As the number of experts increases, so does the demand for computational resources, potentially leading to scalability issues and increased hardware requirements, such as additional GPUs. This could pose a practical constraint, particularly for resource-constrained environments or scenarios where extensive parallel processing capabilities are not readily available.
% While our research shows promising results, it's important to note that the analysis and experiments were conducted solely on the TCGA LUAD dataset. Using a single dataset may limit the generalizability of our approach across diverse datasets and application domains. 

% Another limitation of our study pertains to interpretability. While our approach allows for tracing back to specific data modalities for survival prediction, it does not currently provide the ability to pinpoint the exact biological and clinical factors or features within each modality contributing to the predictions. This lack of granularity in feature attribution within individual modalities restricts our ability to offer detailed insights into the underlying mechanisms driving the model's predictions.

% Another limitation concerns the computational cost associated with adding more experts to the model. As the number of experts increases, so does the demand for computational resources, potentially leading to scalability issues and increased hardware requirements, such as additional GPUs. This could pose a practical constraint, particularly for resource-constrained environments or scenarios where extensive parallel processing capabilities are not readily available. 

\chapter{FUTURE WORK}
This thesis opens several avenues for future research. First, further investigation into the scalability and generalizability of the proposed graph-guided MoE framework across diverse datasets and application domains is warranted. Furthermore, extending the analysis to include additional evaluation metrics and conducting comprehensive comparative studies with existing methods would provide deeper insights into the efficacy of the proposed approach. Moreover, investigating interpretability and explainability techniques for the predictions generated by the graph-guided MoE model could enhance trust and understanding in real-world applications. Additionally, a potential avenue for future research involves developing methods to enhance the interpretability of the graph-guided MoE model specifically for healthcare professionals, such as doctors. Investigating techniques to trace back to the features contributing to survival prediction can provide valuable insights into the biological and clinical factors driving patient outcomes. This could involve developing visualization tools or feature attribution methods that highlight the most influential genomic, transcriptomic, or clinical variables associated with survival. Lastly, exploring techniques to integrate domain knowledge and expert insights into the modeling process could further improve the performance and applicability of the proposed framework in diverse real-world scenarios. Overall, this thesis lays the foundation for future research endeavors aimed at leveraging graph structures and multi-modal data for enhanced predictive modeling and decision-making in complex real-world tasks.

% \begin{table}[htbp]
% \centering
% \caption{Dataset Description}
% \label{tab:dataset-description}
% \small % Decrease font size
% \begin{tabularx}{\linewidth}{|>{\columncolor{gray!30}\centering\arraybackslash}X|>{\centering\arraybackslash}X|}
% \hline
% % \rowcolor{gray!30}
% % \textbf{Attribute} & \textbf{Value} \\
% % \hline
% \cellcolor{gray!30}\textbf{Dataset} & Cancer Genome Atlas TCGA-LUAD (lung adenocarcinoma) cancer dataset \\
% \hline
% \cellcolor{gray!30}\textbf{Data sources} & Genomic Data Commons (GDC) Data Portal \\
% \hline
% \cellcolor{gray!30}\textbf{Total cases} & 585 \\
% \hline
% \cellcolor{gray!30}\textbf{Cases utilized for analysis after merging clinical data} & 470 \\
% \hline
% \cellcolor{gray!30}\textbf{Data includes} & Genomics, diagnostic whole-slide images (WSIs), clinical data \\
% \hline
% \end{tabularx}
% \end{table}

%-----------------------back matter
{\singlespace
% Making the references a "part" rather than a chapter gets it indented at
% level -1 according to the chart: top of page 4 of the document at
% ftp://tug.ctan.org/pub/tex-archive/macros/latex/contrib/tocloft/tocloft.pdf
\addcontentsline{toc}{part}{REFERENCES}
\bibliographystyle{asudis}
\bibliography{dis1}}

@article{[1],
  author = {Chelsea Finn and Pieter Abbeel and Sergey Levine},
  title = {Model-agnostic meta-learning for fast adaptation of deep networks},
  journal = {ICML},
  year = {2017},
}

@article{[Proto],
  author = {Jake Snell and Kevin Swersky and Richard Zemel.},
  title = {Prototypical networks for few-shot learning},
  journal = {NeurIPS},
  year = {2017},
}

@article{[ILP],
  author = {Hirthik Mathavan and Zhen Tan and Nivedh Mudiam and Huan Liu },
  title = {Inductive Linear Probing for Few-Shot Node Classification},
  journal = {SBP-BRiMS},
  year = {2023},
}

@inproceedings{Koch2015SiameseNN,
  title={Siamese Neural Networks for One-Shot Image Recognition},
  author={Gregory R. Koch},
  year={2015}
}

@article{Schroff2015FaceNetAU,
  title={FaceNet: A unified embedding for face recognition and clustering},
  author={Florian Schroff and Dmitry Kalenichenko and James Philbin},
  journal={2015 IEEE Conference on Computer Vision and Pattern Recognition (CVPR)},
  year={2015},
  pages={815-823}
}

@inproceedings{
shazeer2017,
title={ Outrageously Large Neural Networks: The Sparsely-Gated Mixture-of-Experts Layer},
author={Noam Shazeer and *Azalia Mirhoseini and *Krzysztof Maziarz and Andy Davis and Quoc Le and Geoffrey Hinton and Jeff Dean},
booktitle={International Conference on Learning Representations},
year={2017},
}

@article{FSN,
author = {Antoniou, Antreas and Storkey, Amos and Edwards, Harrison},
year = {2017},
month = {11},
pages = {},
title = {Data Augmentation Generative Adversarial Networks}
}

@inproceedings{inproceedings,
author = {Lampert, Christoph and Nickisch, Hannes and Harmeling, Stefan},
year = {2009},
month = {06},
pages = {},
title = {Learning To Detect Unseen Object Classes by Between-Class Attribute Transfer},
journal = {2009 IEEE Computer Society Conference on Computer Vision and Pattern Recognition Workshops, CVPR Workshops 2009},
doi = {10.1109/CVPR.2009.5206594}
}

@inproceedings{NIPS2013_7cce53cf,
 author = {Frome, Andrea and Corrado, Greg S and Shlens, Jon and Bengio, Samy and Dean, Jeff and Ranzato, Marc\textquotesingle Aurelio and Mikolov, Tomas},
 booktitle = {Advances in Neural Information Processing Systems},
 pages = {},
 publisher = {Curran Associates, Inc.},
 title = {DeViSE: A Deep Visual-Semantic Embedding Model},
 volume = {26},
 year = {2013}
}

@inproceedings{NIPS2013_2d6cc4b2,
 author = {Socher, Richard and Ganjoo, Milind and Manning, Christopher D and Ng, Andrew},
 booktitle = {Advances in Neural Information Processing Systems},
 pages = {},
 publisher = {Curran Associates, Inc.},
 title = {Zero-Shot Learning Through Cross-Modal Transfer},
 volume = {26},
 year = {2013}
}

@ARTICLE{10061470,
  author={Hou, Wentai and Lin, Chengxuan and Yu, Lequan and Qin, Jing and Yu, Rongshan and Wang, Liansheng},
  journal={IEEE Transactions on Medical Imaging}, 
  title={Hybrid Graph Convolutional Network With Online Masked Autoencoder for Robust Multimodal Cancer Survival Prediction}, 
  year={2023},
  volume={42},
  number={8},
  pages={2462-2473}}

@INPROCEEDINGS{10204656,
  author={Kang, Mingu and Song, Heon and Park, Seonwook and Yoo, Donggeun and Pereira, Sérgio},
  booktitle={2023 IEEE/CVF Conference on Computer Vision and Pattern Recognition (CVPR)}, 
  title={Benchmarking Self-Supervised Learning on Diverse Pathology Datasets}, 
  year={2023},
  volume={},
  number={},
  pages={3344-3354},
  doi={10.1109/CVPR52729.2023.00326}}

@article{WANG2022102559,
title = {Transformer-based unsupervised contrastive learning for histopathological image classification},
journal = {Medical Image Analysis},
volume = {81},
pages = {102559},
year = {2022},
issn = {1361-8415},
author = {Xiyue Wang and Sen Yang and Jun Zhang and Minghui Wang and Jing Zhang and Wei Yang and Junzhou Huang and Xiao Han},
}

@inproceedings{AVIS,
  title={AVIS: Autonomous Visual Information Seeking with Large Language Model Agent.},
  author={Ziniu Hu and Ahmet Iscen and Chen Sun and Kai-Wei Chang and Yizhou Sun and David A Ross and Cordelia Schmid and Alireza Fathi},
  year={2023}
}

@inproceedings{SociS,
  title={Dynamic Few-shot Learning for Computational Social Science.},
  author={Malla R and Coan T and Srinivasan V and Boussalis C},
  year={2024}
}

@inproceedings{CyberS,
author = {Sureshan and Shruti and Das and Debasis},
title = {Few-Shot Learning based Anomaly Detection in Security Applications},
year = {2023},
isbn = {9781450397971},
publisher = {Association for Computing Machinery},
address = {New York, NY, USA},
doi = {10.1145/3570991.3571040},
booktitle = {Proceedings of the 6th Joint International Conference on Data Science \& Management of Data (10th ACM IKDD CODS and 28th COMAD)},
pages = {295–296},
numpages = {2},
location = {Mumbai, India},
series = {CODS-COMAD '23}
}

@article{GE2023104458,
title = {Few-shot learning for medical text: A review of advances, trends, and opportunities},
journal = {Journal of Biomedical Informatics},
volume = {144},
pages = {104458},
year = {2023},
issn = {1532-0464},
doi = {https://doi.org/10.1016/j.jbi.2023.104458},
author = {Yao Ge and Yuting Guo and Sudeshna Das and Mohammed Ali Al-Garadi and Abeed Sarker}
}

@article{1Bray2018,
  title={Global cancer statistics 2018: GLOBOCAN estimates of incidence and mortality worldwide for 36 cancers in 185 countries},
  author={Bray, Freddie and Ferlay, Jacques and Soerjomataram, Isabelle and Siegel, Rebecca L. and Torre, Lindsey A. and Jemal, Ahmedin},
  journal={CA: A Cancer Journal for Clinicians},
  volume={68},
  pages={394--424},
  year={2018},
  doi={10.3322/caac.21492}
}

@article{2Mariotto2014,
  title={Cancer survival: An overview of measures, uses, and interpretation},
  author={Mariotto, Angela B. and Noone, Anne-Michelle and Howlader, Nadia and Cho, Hyunsoon and Keel, Gretchen E. and Garshell, Jessica and Woloshin, Steven and Schwartz, Lisa M.},
  journal={Journal of the National Cancer Institute Monographs},
  pages={145--186},
  year={2014},
  doi={10.1093/jncimonographs/lgu024}
}

@article{3Simmons2017,
  title={Prognostic tools in patients with advanced cancer: A systematic review},
  author={Simmons, Charlene P. L. and McMillan, Donald C. and McWilliams, Kelly and Sande, Terence A. and Fearon, Ken C. H. and Tuck, Scott and Fallon, Marie T.},
  journal={Journal of Pain and Symptom Management},
  volume={53},
  pages={962--970},
  year={2017},
  doi={10.1016/j.jpainsymman.2016.12.330}
}

@article{4Hui2019,
  title={Prognostication in advanced cancer: Update and directions for future research},
  author={Hui, David and Hess, Kenneth and dos Santos, Renata and Chisholm, Gary and Bruera, Eduardo},
  journal={Supportive Care in Cancer},
  volume={27},
  pages={1973--1984},
  year={2019},
  doi={10.1007/s00520-019-04727-y}
}

@article{5Cheon2016,
  title={The accuracy of clinicians predictions of survival in advanced cancer: A review},
  author={Cheon, Sarah and Agarwal, Amita and Popovic, Marko and Milakovic, Momcilo and Lam, Maggie and Chan, Sarah and Krzyzanowska, Monika K.},
  journal={Annals of Palliative Medicine},
  volume={5},
  pages={22--29},
  year={2016},
  doi={10.3978/j.issn.2224-5820.2015.08.04}
}

@article{6Cox1972,
  title={Regression models and life-tables},
  author={Cox, David R.},
  journal={Journal of the Royal Statistical Society: Series B (Methodological)},
  volume={34},
  pages={187--220},
  year={1972},
  doi={10.1111/j.2517-6161.1972.tb00899.x}
}

@article{7LeCun2015,
  title={Deep learning},
  author={LeCun, Yann and Bengio, Yoshua and Hinton, Geoffrey},
  journal={Nature},
  volume={521},
  pages={436--444},
  year={2015},
  doi={10.1038/nature14539}
}

@article{10Norgeot2019,
  title={A call for deep-learning healthcare},
  author={Norgeot, Benjamin and Glicksberg, Benjamin S. and Butte, Atul J.},
  journal={Nature Medicine},
  volume={25},
  pages={14--15},
  year={2019},
  doi={10.1038/s41591-018-0320-3}
}

@article{12Katzman2018,
  title={DeepSurv: Personalized treatment recommender system using a Cox proportional hazards deep neural network},
  author={Katzman, Jared L. and Shaham, Uri and Cloninger, Alexander and Bates, Jonathan and Jiang, Tingting and Kluger, Yuval},
  journal={BMC Medical Research Methodology},
  year={2018},
  doi={10.1186/s12874-018-0482-1}
}

@article{13Ching2018,
  title={Cox-nnet: An artificial neural network method for prognosis prediction of high-throughput omics data},
  author={Ching, Travers and Zhu, Xiang and Garmire, Lana X.},
  journal={PLoS Computational Biology},
  volume={14},
  pages={e1006076},
  year={2018},
  doi={10.1371/journal.pcbi.1006076}
}

@article{14Lu2019,
  title={Deep learning to assess long-term mortality from chest radiographs},
  author={Lu, M. T. and et al.},
  journal={JAMA Netw. Open},
  volume={2},
  pages={e197416},
  year={2019},
  doi={10.1001/jamanetworkopen.2019.7416}
}

@article{15Mukherjee2020,
  title={A shallow convolutional neural network predicts prognosis of lung cancer patients in multi-institutional CT-image data},
  author={Mukherjee, P. and et al.},
  journal={Nat. Mach. Intell.},
  volume={2},
  pages={274–282},
  year={2020},
  doi={10.1038/s42256-020-0173-6}
}

@article{16Zhang2020,
  title={A deep learning risk prediction model for overall survival in patients with gastric cancer: A multicenter study},
  author={Zhang, L. and et al.},
  journal={Radiother. Oncol.},
  volume={150},
  pages={73–80},
  year={2020},
  doi={10.1016/j.radonc.2020.06.010}
}

@article{17Zhong2020,
  title={A deep learning MR-based radiomic nomogram may predict survival for nasopharyngeal carcinoma patients with stage T3N1M0},
  author={Zhong, L.-Z. and et al.},
  journal={Radiother. Oncol.},
  volume={151},
  pages={1–9},
  year={2020},
  doi={10.1016/j.radonc.2020.06.050}
}

@article{20Baltrusaitis2019,
  title={Multimodal machine learning: A survey and taxonomy},
  author={Baltrusaitis, T. and Ahuja, C. and Morency, L.-P.},
  journal={IEEE T. Pattern Anal.},
  volume={41},
  pages={423–443},
  year={2019},
  doi={10.1109/TPAMI.2018.2798607}
}

@article{21Yousefi2017,
  title={Predicting clinical outcomes from large scale cancer genomic profiles with deep survival models},
  author={Yousefi, S. and et al.},
  journal={Sci. Rep.},
  volume={7},
  pages={11707},
  year={2017},
  doi={10.1038/s41598-017-11817-6}
}

@article{22Mobadersany2018,
  title={Predicting cancer outcomes from histology and genomics using convolutional networks},
  author={Mobadersany, P. and et al.},
  journal={Proc. Natl. Acad. Sci. U.S.A.},
  volume={115},
  pages={E2970–E2979},
  year={2018},
  doi={10.1073/pnas.1717139115}
}

@article{23Huang2019,
  title={SALMON: Survival analysis learning with multi-omics neural networks on breast cancer},
  author={Huang, Z. and et al.},
  journal={Front. Genet.},
  volume={10},
  pages={166},
  year={2019},
  doi={10.3389/fgene.2019.00166}
}

@article{24Cheerla2019,
  title={Deep learning with multimodal representation for pancancer prognosis prediction},
  author={Cheerla, A. and Gevaert, O.},
  journal={Bioinformatics},
  volume={35},
  pages={i446–i454},
  year={2019},
  doi={10.1093/bioinformatics/btz342}
}

@inproceedings{SNA,
author = {Liu, Pengyuan and De Sabbata, Stef},
year = {2019},
month = {11},
pages = {},
title = {Estimating Locations of Social Media Content through a Graph-based Link Prediction},
isbn = {978-1-4503-7260-2},
journal = {GIR '19: Proceedings of the 13th Workshop on Geographic Information Retrieval},
doi = {10.1145/3371140.3371141}
}

@inproceedings{ReSs,
author = {Ying, Rex and He, Ruining and Chen, Kaifeng and Eksombatchai, Pong and Hamilton, William L. and Leskovec, Jure},
title = {Graph Convolutional Neural Networks for Web-Scale Recommender Systems},
year = {2018},
isbn = {9781450355520},
publisher = {Association for Computing Machinery},
address = {New York, NY, USA},
doi = {10.1145/3219819.3219890},
booktitle = {Proceedings of the 24th ACM SIGKDD International Conference on Knowledge Discovery \& Data Mining},
pages = {974–983},
numpages = {10},
location = {London, United Kingdom},
series = {KDD '18}
}

@inproceedings{BINF,
author = {Yi HC and You ZH and Huang DS and Kwoh CK.},
year = {2019},
month = {11},
pages = {},
title = {Graph representation learning in bioinformatics: trends, methods and applications.},
isbn = {978-1-4503-7260-2},
journal = {GIR '19: Proceedings of the 13th Workshop on Geographic Information Retrieval},
doi = {10.1093/bib/bbab340.}
}

@string{nature = {Nature}}

@string{science = {Science}}
% \renewcommand{\chaptername}{APPENDIX}
% \addtocontents{toc}{APPENDIX \par}
% \appendix
% \include{appendix1}
\include{vita}
\end{document}